\documentclass[10pt,twocolumn,letterpaper]{article}

\usepackage[pagenumbers]{cvpr} % To force page numbers, e.g. for an arXiv version
\usepackage{hyperref}
\usepackage{url}
\usepackage{graphicx}
\usepackage{wrapfig}
\usepackage{booktabs}       % professional-quality tables
\usepackage{multirow} 
\usepackage{amssymb}
\usepackage{pifont}
\usepackage{algorithm}
\usepackage{algorithmic}
\usepackage{pifont}
\usepackage{bm}
\usepackage{cuted}      % 提供 strip 环境
\usepackage{caption}    % 更好控制 caption 样式

\usepackage{color} % (if used in text)

\usepackage{xcolor}
\usepackage{colortbl,booktabs}

\definecolor{citecolor}{RGB}{17,80,197}
\hypersetup{
    colorlinks=true,
    linkcolor=red,
    citecolor=citecolor,
    filecolor=magenta,      
    urlcolor=magenta,
}
\definecolor{cvprblue}{rgb}{0.21,0.49,0.74}

\def\confName{CVPR}
\def\confYear{2027}

\title{WorldUAV: Scaling UAV Data Synthesis via World Models}

\author{Xuewen Liu$^{1,2}$, \; Mengjuan Chen$^{1}$, \; Jing Zhang$^{1,2}$, \; Mengzhe Shen$^{1}$, \; Zekun Xu$^{1}$, \; Zibo Chen$^{1}$, \\
\vspace{0.2cm}
Zuoxiu Zhang$^{1}$, \; Yue Dong$^{1}$, \; Zhikai Li$^{1}$\thanks{Corresponding author: \{zhikai.li, qingyi.gu\}@ia.ac.cn.}, \; Qingyi Gu$^{1}$\footnotemark[1]  \\
$^1$Institute of Automation, Chinese Academy of Sciences\\
$^2$School of Artificial Intelligence, University of Chinese Academy of Sciences\\
\vspace{0.2cm}
{\tt\small \{liuxuewen2023, zhikai.li, qingyi.gu\}@ia.ac.cn} \\
{\tt\small \textbf{Project Page:} \url{https://bienluky.github.io/WorldUAV/}}
}

\begin{document}
\maketitle

\begin{abstract}
UAV detection and tracking remain constrained by the scarcity of high-quality training data.
Existing game-engine-based synthetic UAV data suffer from a substantial synthetic-to-real gap, while the high cost of manually collected real-world UAV data limits dataset scale.
Therefore, we introduce WorldUAV, a high-quality UAV dataset constructed through an efficient framework powered by world models.
Specifically, WorldUAV directly integrates UAVs into readily available real-world videos, naturally preserving realistic backgrounds, illumination, and scene dynamics. Leveraging the spatiotemporal priors of world models, it generates UAV videos with plausible motion, pose variations, and cross-frame consistency, effectively narrowing the gap between synthetic and real data.
We further develop an agentic pipeline orchestrated by a multimodal large language model that acquires sources, plans UAV trajectories, invokes synthesis, and verifies generated videos, enabling fully automated dataset construction.
WorldUAV comprises 6,878 UAV videos and 205,778 UAV images.
Experiments show that WorldUAV significantly improves the performance of UAV detectors. 
Moreover, it substantially reduces dataset construction costs, requiring only 15 seconds and USD 0.0005 API cost per high-quality UAV image.
Our work offers both a high-quality dataset and a scalable data generation paradigm for UAV detection and tracking.
\end{abstract}    
\section{Introduction}
As Unmanned Aerial Vehicles (UAVs) become increasingly prevalent across diverse applications, their growing flight activities also pose new challenges to public safety, low-altitude airspace management, and critical infrastructure protection~\cite{dafrallah2024malicious}.
UAV detection and tracking, as key technologies for counter-UAV defense, have attracted increasing attention.
Recent advances in deep learning have substantially improved their performance.
However, further progress remains heavily dependent on high-quality training data, which existing UAV datasets struggle to provide.

Beyond scale, a high-quality UAV dataset should capture the complexity of real-world conditions. 
This requires broad coverage of diverse scenes, weather and lighting conditions, UAV models, and flight poses, while maintaining visual fidelity to real-world observations~\cite{lenhard2025syndronevision}. 
For tracking, continuous video sequences are equally essential to capture target motion and temporal dynamics~\cite{shim2025focusing}. 
Therefore, balancing scale, diversity, realism, and temporal richness is crucial for constructing high-quality UAV datasets.

Existing UAV datasets primarily rely on real-world collection~\cite{zhao2022vision} or simulation-based synthesis~\cite{lenhard2025syndronevision}.
Real-world collection preserves authentic visual characteristics and temporal dynamics, but high acquisition and annotation costs, coupled with limited control over environmental and flight conditions, constrain its scalability and diversity.
In contrast, synthetic approaches offer flexible control over scenes and UAV attributes, enabling scalable data generation.
SynDroneVision~\cite{lenhard2025syndronevision} synthesizes UAV images by simulating real-world scenes with a game engine. 
However, discrepancies in background textures, illumination, and imaging characteristics introduce a substantial synthetic-to-real gap.
For example, despite comprising over 140K images across diverse scenes, SynDroneVision-trained YOLOv8m achieves only 0.639 mAP@0.5 on the real-world DUT Anti-UAV (DUT)~\cite{zhao2022vision} test set, substantially below the 0.933 achieved with DUT-only training.
Moreover, existing datasets predominantly focus on image-level detection, with limited support for UAV tracking. 
For instance, DUT contains only 20 UAV tracking videos.

\begin{figure*}[t]
    \centering
    \includegraphics[width=\textwidth]{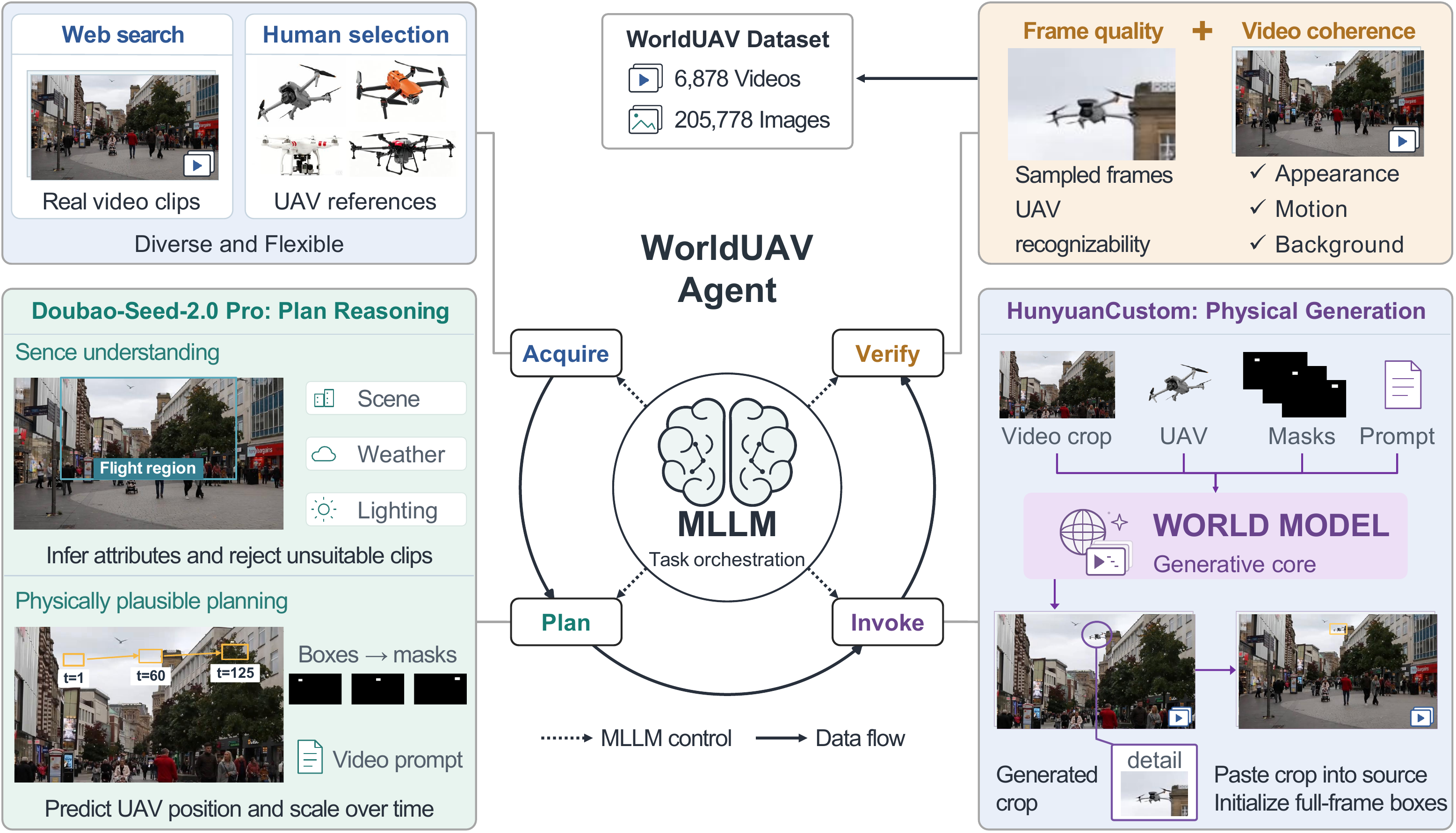}
    \caption{\textbf{Overview of WorldUAV Agent.} An MLLM orchestrates tool use, scene-aware planning, and quality verification, with HunyuanCustom serving as the world-model generative core.
    Retrieved scene videos and manually selected UAV references support diverse and flexible synthesis.
    The MLLM understands the video scene and predicts UAV positions and scales over time.
    The agent invokes the world model with multimodal guidance to generate physically plausible UAV video crops, and initializes full-frame bounding boxes.
    MLLM-based checks retain videos with recognizable UAVs and temporal coherence.}
    \label{fig:construction_pipeline}
\end{figure*}

World models~\cite{hu2025hunyuancustom} offer a promising alternative to conventional UAV data synthesis through their powerful spatiotemporal modeling and controllable video generation capabilities. 
Unlike game-engine simulations that construct virtual environments, world models with video editing capabilities can directly integrate UAVs into real-world videos, preserving authentic backgrounds, illumination, and scene dynamics. 
Leveraging learned spatiotemporal priors, they further enable plausible UAV trajectories, pose variations, and temporal consistency. 
By combining real-world visual realism with controllable synthesis, this paradigm offers a scalable approach to UAV video generation while mitigating the distribution gap between synthetic and real data.

Building on these insights, we introduce WorldUAV, a synthetic UAV dataset featuring large scale, diverse coverage, high realism, and rich temporal information.
The core of WorldUAV is to leverage world models for controllable UAV synthesis in real-world videos. 
This enables seamless UAV integration while preserving scene realism and temporal consistency.
To enable efficient and scalable dataset construction, we develop WorldUAV Agent, an agentic framework orchestrated by a multimodal large language model (MLLM). 
As shown in Fig.~\ref{fig:construction_pipeline}, the MLLM uses web search tools to retrieve real-world scene videos, while UAV types are selected manually. 
For each retrieved video, it reasons over scene semantics and spatial layouts to identify suitable flight regions, plan UAV trajectories as spatiotemporal masks, and derive video attributes and synthesis prompts.
With reference images, trajectory masks, and prompts as conditioning inputs, the agent invokes the world model to synthesize physically plausible UAVs within selected video crops. 
Generated crops are reintegrated into source videos, with bounding box annotations updated for all frames.
MLLM-based verification assesses UAV recognizability and video coherence, while a fixed detector selects image frames from accepted videos. 
The pipeline also supports flexible data customization by specifying target scenes or UAV types.

Using this pipeline, we construct WorldUAV, comprising 6,878 UAV videos and 205,778 images, covering 7 scene categories, 5 weather conditions, 4 lighting, 12 UAV models, and 8 flight viewpoints. 
Extensive experiments demonstrate that WorldUAV consistently improves UAV detection performance and generalization across diverse datasets.
YOLOv12l trained on WorldUAV with DUT achieves 0.957 mAP@0.5 on the DUT test set and 0.837 on the out-of-distribution UAV-Eagle~\cite{barivsic2022brain} dataset.
Replacing the detector in existing tracking models with one trained on WorldUAV also improves UAV tracking performance.
Moreover, our pipeline efficiently synthesizes high-quality UAV images in just 15 seconds per image, with an API cost of only USD 0.0005 each.
Overall, our key contributions are as follows:
\begin{itemize}
    \item We introduce a novel world-model-driven paradigm for UAV data synthesis that integrates UAVs into real-world videos, mitigating the synthetic-to-real gap.
    \item We construct WorldUAV, a high-quality synthetic UAV dataset that substantially enhances UAV detection, tracking, and cross-dataset generalization.
    \item We develop an MLLM-orchestrated agentic framework, enabling efficient, scalable, and customizable UAV video generation with quality control.
\end{itemize}

\section{Related Work}
\label{sec:related_work}

\subsection{Datasets for UAV Detection and Tracking}
\label{sec:related_uav_datasets}

High-quality UAV datasets are essential for advancing UAV detection and tracking.
Existing datasets are primarily constructed through real-world collection or simulation-based synthesis.
DUT Anti-UAV~\cite{zhao2022vision} provides annotated RGB images and videos for detection and short- and long-term tracking.
Anti-UAV~\cite{jiang2021anti} offers paired RGB and infrared tracking sequences with frame-level bounding boxes.
RealDroneVision~\cite{sivapuram2026realdronevision} targets small-drone detection with video-derived images and semi-automatic annotation supported by human verification.
UAV-CB~\cite{huang2026uav} focuses on camouflaged UAVs in cluttered backgrounds using paired visible and thermal images.
Although real-world collection preserves authentic visual characteristics, high acquisition and annotation costs limit dataset scale and diversity.
Synthetic datasets enable cost-effective data generation through controllable variations in scenes and UAV attributes.
Sim2Air~\cite{barisic2022sim2air} renders UAVs in Blender with randomized textures to encourage shape-based recognition under challenging illumination and small target sizes.
SynDroneVision~\cite{lenhard2025syndronevision} uses game-engine simulation to generate RGB images with diverse backgrounds, lighting conditions, and drone models.
However, differences between rendered and real imagery can limit the transfer of models trained on synthetic data to real-world observations.
Therefore, constructing UAV datasets that jointly achieve visual realism, scalability, and diversity remains challenging.

\subsection{Data Synthesis}
\label{sec:related_synthetic_data}

Synthetic visual data can be constructed through simulation, image compositing, and learned generation.
Simulation provides control over scene content and access to ground-truth annotations.
SYNTHIA~\cite{ros2016synthia} renders urban scenes with semantic labels, while Kubric~\cite{greff2022kubric} combines Blender rendering with PyBullet physics to generate annotated videos.
Image compositing reuses existing imagery: SynthText~\cite{gupta2016synthetic} places text into natural images using scene geometry, and Copy-Paste~\cite{ghiasi2021simple} combines objects from different images for instance segmentation.
Diffusion models further enable data synthesis from text and spatial conditions.
% StableRep~\cite{tian2023stablerep} treats images generated from the same prompt as positive samples for contrastive learning.
Dataset Diffusion~\cite{nguyen2023dataset} derives segmentation pseudolabels from diffusion attention maps.
Extending synthesis to temporal data, TrackDiffusion~\cite{li2025trackdiffusion} generates consistent videos from object tracklets for tracker training.
Generative world models, such as Cosmos~\cite{nvidia2025cosmos}, learn spatiotemporal priors from large-scale videos to model scene dynamics under conditioning inputs.
These capabilities offer a promising basis for synthesizing diverse video data with plausible motion and temporal consistency.
Building on this potential, we use world models to integrate UAVs into real-world videos for detection and tracking.

% Reproduces Table 1 of the user-supplied SynDroneVision.pdf (arXiv v1,
% printed p. 3): original row order, values, symbols, grid, and footnotes.
% Extensions: UAV Video column, RealDroneVision row, and blank WorldUAV row.
% References use this manuscript's bibliography numbering rather than hard-coded
% numbers from SynDroneVision. Evidence: tab/dataset_comparison_sources.md.
% Requires graphicx, pifont, amssymb, and citation support (already in main.tex).
% Insert where needed with: \input{tab/dataset_comparison}
\begin{table*}[t]
    \centering
    \caption{Comprehensive information on publicly available datasets for image-based drone detection, encompassing both real and synthetic data.
    Original entries follow Table~1 of SynDroneVision~\cite{lenhard2025syndronevision}.
    The symbols \ding{55} and \ding{51} indicate absence and presence, respectively; Img. Sequ., Diff. Drones, and Dist. Objs. denote image sequences, diverse drone models, and distractor objects.}
    \label{tab:dataset_comparison}
    \begingroup
    \small
    \setlength{\tabcolsep}{3pt}
    \renewcommand{\arraystretch}{1.05}
    % Use black symbols, independently of the colored cmark/xmark macros in main.tex.
    \newcommand{\dtYes}{\ding{51}}
    \newcommand{\dtNo}{\ding{55}}
    \resizebox{0.9\textwidth}{!}{%
    \begin{tabular}{ l|r r r r|c c c | r}
        \hline
        Dataset & \multicolumn{4}{c|}{No. Images} & Img. & Diff. & Dist. & Max. Img. \\
        & \multicolumn{1}{c}{train} & \multicolumn{1}{c}{val} & \multicolumn{1}{c}{test} & \multicolumn{1}{c|}{total} & Sequ. & Drones & Objs. & Resolution \\
        \hline
        USC Drone Detect. \& Track.~\cite{chen2017deep,wang2019towards} & -- & -- & -- & 27,000\textsuperscript{$\bigstar$} & \dtYes & \dtNo & \dtNo & $1920\!\times\!1080$ \\
        Drone Dataset~\cite{aksoy2019drone} & 3,611 & -- & 401 & 4,012 & \dtNo & \dtNo & \dtNo & $3840\!\times\!2160$ \\
        MAV-VID~\cite{rodriguez2020adaptive} & 29,500 & 10,732 & -- & 40,232 & \dtYes & \dtYes & \dtNo & -- \\
        Det-Fly~\cite{zheng2021air} & -- & -- & -- & 13,271\textsuperscript{$\bigstar$} & \dtYes & \dtNo & \dtNo & $3840\!\times\!2160$ \\
        UAV-Eagle~\cite{barivsic2022brain} & -- & -- & -- & 510\textsuperscript{$\bigstar$} & \dtYes & \dtNo & \dtNo & $1920\!\times\!1080$ \\
        UAVData~\cite{zeng2021uavdata} & -- & -- & -- & 13,803\textsuperscript{$\bigstar$} & \dtYes & \dtYes & \dtYes & $1280\!\times\!720$ \\
        Halmstadt Data~\cite{svanstrom2021dataset} & -- & -- & -- & 203,328\textsuperscript{$\blacktriangle$} & \dtYes & \dtYes & \dtYes & $640\!\times\!512$ \\
        DUT Anti-UAV~\cite{zhao2022vision} & 5,200 & 2,600 & 2,200 & 10,000 & \dtYes & \dtYes & \dtNo & $5616\!\times\!3744$ \\
        VisioDECT~\cite{sakib2025visiodect} & -- & -- & -- & 20,924\textsuperscript{$\bigstar$} & \dtYes & \dtYes & \dtNo & $852\!\times\!480$ \\
        Malicious Drones~\cite{jamil2022distinguishing} & 543 & -- & 233 & 776 & \dtNo & \dtYes & \dtYes & $224\!\times\!224$ \\
        S-UAV-T~\cite{barisic2022sim2air} (\textit{synthetic}) & -- & -- & -- & 52,500\textsuperscript{$\bigstar$} & \dtNo & \dtYes & \dtNo & $608\!\times\!608$ \\
        Drone-vs-Bird Detection Ch.~\cite{coluccia2024drone} & 85,904 & -- & --\textsuperscript{$\ast$} & 85,904 & \dtYes & \dtYes & \dtYes & $3840\!\times\!2160$ \\
        SynDroneVision~\cite{lenhard2025syndronevision} (\textit{synthetic}) & 131,238 & 8,800 & 4,000 & 140,038 & \dtYes & \dtYes & \dtYes & $2560\!\times\!1489$ \\
        RealDroneVision~\cite{sivapuram2026realdronevision} & 138,417 & -- & 34,606 & 173,023 & \dtYes & \dtYes & \dtYes & $1920\!\times\!1080$ \\
        \hline
        \bf WorldUAV (\textit{synthetic}) & \bf 190,778 & \bf 5,000 & \bf 10,000 & \bf 205,778 & \dtYes & \dtYes & \dtYes & $\bm{4096\!\times\!2160}$\\
        \hline
    \end{tabular}%
    }
    \par\vspace{2pt}
    \begin{minipage}{0.9\textwidth}
        \footnotesize
        $\ast$ not publicly available \quad
        $\bigstar$ no subdivision into train, val, and test \quad
        $\blacktriangle$ RGB + IR data.
    \end{minipage}
    \endgroup
    \vspace{-0.3cm}
\end{table*}

\section{WorldUAV Dataset}
\label{sec:construction}

\subsection{Construction Pipeline}
\label{sec:construction_pipeline}

WorldUAV Agent is an MLLM-orchestrated framework with a world model as its generative core, as shown in Fig.~\ref{fig:construction_pipeline}.
We use Doubao-Seed-2.0 Pro~\cite{bytedance2026seed2} as the MLLM and find it sufficient for task orchestration in our experiments; the framework also supports alternative MLLMs.
The MLLM retrieves real-world videos using web search tools, while UAV types and reference images are selected manually.
Users may also supply source videos, enabling flexible synthesis tailored to specific scenes and UAV types.
All source videos are resampled to a uniform frame rate of 25 fps and segmented into 125-frame clips matching the input length of world model.

The MLLM infers scene, weather, and lighting attributes from sampled frames, which are retained as dataset annotations. 
Guided by scene understanding, it selects an $832 \times 480$ region with suitable flight space to match the world model's input resolution. 
Clips without a suitable UAV region are excluded. 
For each selected crop, the MLLM generates a synthesis prompt and per-frame normalized bounding boxes to guide physically plausible UAV motion.
Each box specifies the intended UAV center and size.
The planning instructions account for scene layout and camera motion to encourage smooth trajectories and gradual scale changes. 
The agent rasterizes these boxes into binary masks that encode the UAV trajectory.

Given the cropped video, a UAV reference image, the mask sequence, and the synthesis prompt, the agent invokes the HunyuanCustom~\cite{hu2025hunyuancustom} as the world model.
The reference guides UAV appearance, while the masks specify the intended position and scale. 
The world model performs physically plausible generation guided by its learned spatiotemporal priors.
The generated crop is composited back into the source frames, preserving the surrounding scene outside the edited region.
Planned boxes are mapped from crop coordinates to full-frame coordinates to initialize annotations for every frame.
Notably, the strong generative capabilities of the world model enable precise UAV localization within the mask, ensuring reliable initial bounding-box annotations.

Finally, MLLM evaluates synthesized videos for both target recognizability and temporal coherence.
For frame-level assessment, it inspects local crops around the planned UAV positions in 20 uniformly sampled frames, requiring the UAV to be recognizable in more than 75\% of them.
Video-level assessment considers appearance consistency, motion continuity, and background stability, rejecting clips with identity changes, abrupt motion or scale jumps, or severe flickering.
Only clips passing both checks enter the WorldUAV video dataset.
Each video frame can serve as a UAV image.
To ensure a standardized and reproducible video-to-image conversion, we adopt a fixed detector-based selection criterion. Specifically, a trained YOLOv12m detector selects frames whose detected UAV boxes overlap the initialized annotations with IoU greater than 0.75.
Each retained image is paired with its bounding box and the UAV model, viewpoint, scene, weather, and lighting metadata.

\subsection{Dataset Diversity}
\label{sec:dataset_coverage}

WorldUAV draws on real-world footage organized under 2,677 source-video identifiers in the video collection.
Diversity arises from the backgrounds and environmental conditions of this footage, together with the UAV reference appearances used for synthesis.
We describe the dataset attribute vocabulary in this section.

\begin{figure}[t]
    \centering
    \includegraphics[width=\columnwidth]{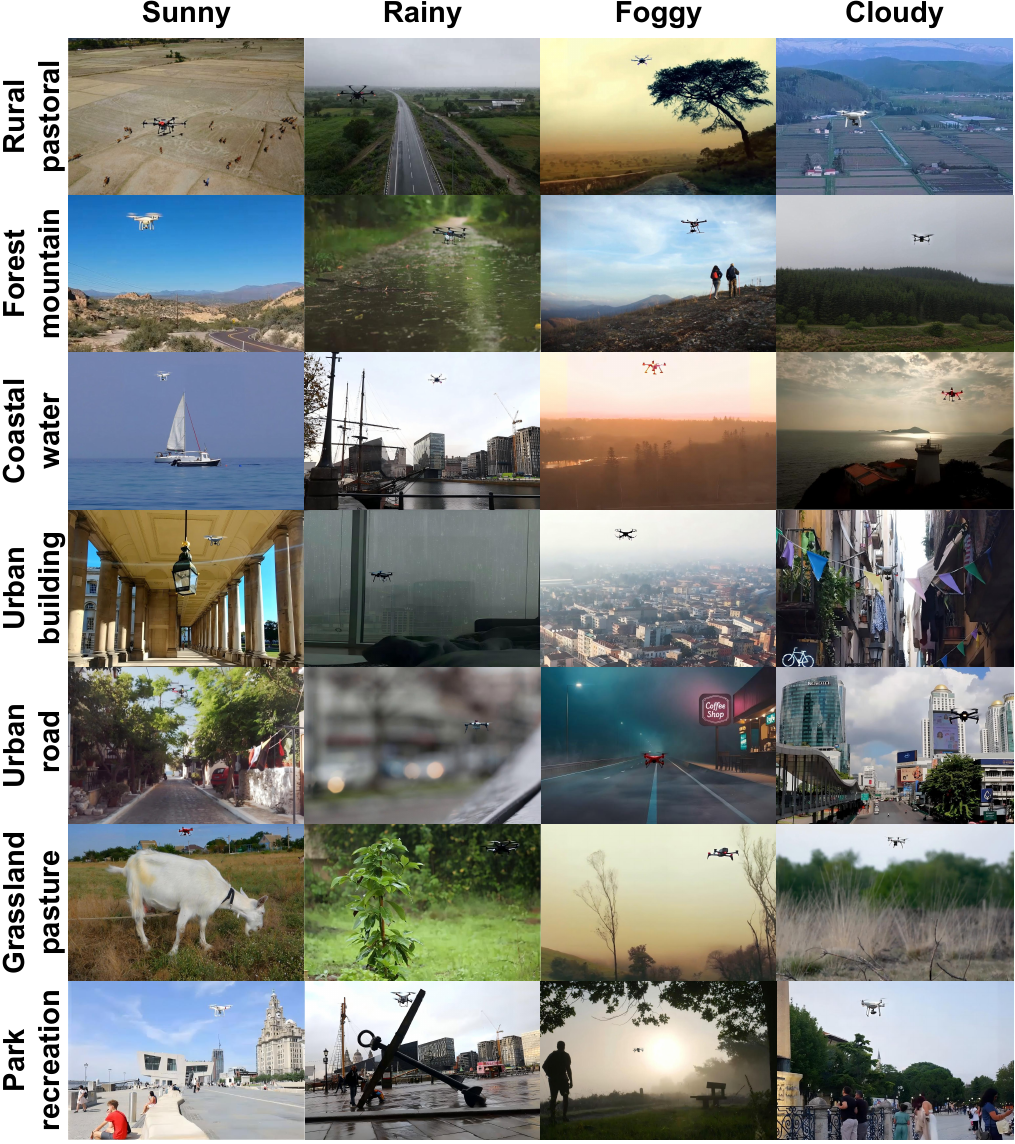}
    \vspace{-0.2cm}
    \caption{\textbf{Visualization of WorldUAV images.} The examples show the diversity of backgrounds and environmental conditions.}
    \label{fig:visualization_sence}
    \vspace{-0.3cm}
\end{figure}

\vspace{0.2cm}
\noindent
\textbf{Background Scenes.}
As shown in Fig.~\ref{fig:visualization_sence}, the scene vocabulary comprises seven categories: \emph{rural/pastoral}, \emph{urban road}, \emph{urban building}, \emph{park/recreation}, \emph{coastal/water}, \emph{grassland/pasture}, and \emph{forest/mountain}.
These categories encompass farmland and villages, streets and traffic, building facades and skylines, recreational spaces, shorelines and waterways, open grasslands, and wooded or mountainous terrain, respectively.
They provide backgrounds with different spatial structures, from relatively open sky and water regions to cluttered vegetation and built environments.
% Each clip and image receives a label indicating its dominant scene.

\vspace{0.2cm}
\noindent 
\textbf{Lighting and Weather.}
We distinguish four lighting conditions, \emph{dawn}, \emph{daylight}, \emph{dusk}, and \emph{night}, from five weather conditions, \emph{sunny}, \emph{cloudy}, \emph{foggy}, \emph{rainy}, and \emph{snowy}.
Lighting describes scene illumination conditions, while weather captures atmospheric conditions.
These attributes are automatically inferred from the original source footage by the MLLM and inherited by the synthesized samples.
Consequently, they reflect observed environmental conditions rather than independently adjustable rendering parameters.
% s; their joint distribution is reported in Fig.~\ref{fig:worlduav_attribute_coverage}.

% \begin{figure}[t]
%     \centering
%     \includegraphics[width=\columnwidth]{image/worlduav_attribute_coverage.pdf}
%     \caption{\textbf{Joint environmental coverage.} Each cell reports a percentage of the 190,778 training images, counted once per image. Scene and weather labels are inherited from video metadata; frequencies reflect the observed collection rather than a balanced combination of attributes.}
%     \label{fig:worlduav_attribute_coverage}
% \end{figure}

\begin{figure}[t]
    \centering
    \includegraphics[width=\columnwidth]{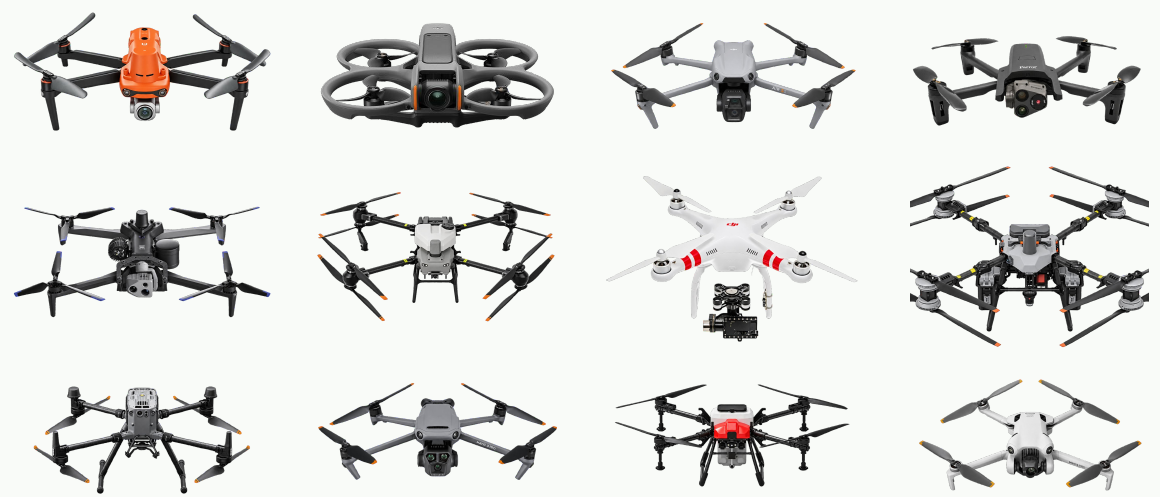}
    \vspace{-0.4cm}
    \caption{\textbf{Visualization of UAV models.} We select 12 UAV models with diverse sizes and appearances for data synthesis.}
    \label{fig:visualization_uav}
    \vspace{-0.2cm}
\end{figure}

\begin{figure}[t]
    \centering
    \includegraphics[width=\columnwidth]{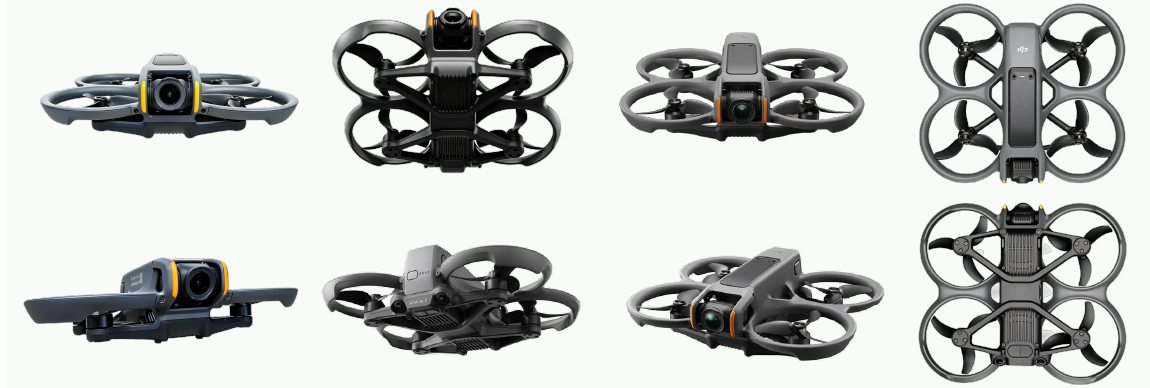}
    \vspace{-0.4cm}
    \caption{\textbf{Visualization of UAV viewpoints.} The eight categories provide comprehensive coverage of UAV viewing angles.}
    \label{fig:visualization_viewpoint}
    \vspace{-0.2cm}
\end{figure}

\begin{table}[t]
    \centering
    \caption{\textbf{Composition of WorldUAV.} Videos and Images report the numbers of video clips and selected training images in each background scene. Lighting, Weather, UAV Model, and Viewpoint report the numbers of distinct categories observed in the corresponding video metadata.}
    \vspace{-0.2cm}
    \label{tab:worlduav_composition}
    \Huge
    \resizebox{\columnwidth}{!}{%
    \begin{tabular}{lrrrrrr}
        \toprule
        Scene & Videos & Images & Lighting & Weather & Model & Viewpoint \\
        \midrule
        Rural/pastoral & 1,833 & 58,119 & 4 & 5 & 4 & 7 \\
        Forest/mountain & 1,440 & 36,290 & 4 & 5 & 8 & 7 \\
        Coastal/water & 1,150 & 27,136 & 4 & 5 & 8 & 7 \\
        Urban building & 922 & 25,390 & 4 & 5 & 8 & 7 \\
        Urban road & 613 & 16,309 & 4 & 5 & 8 & 6 \\
        Grassland/pasture & 465 & 15,915 & 3 & 4 & 4 & 5 \\
        Park/recreation & 455 & 11,619 & 4 & 5 & 8 & 7 \\
        \midrule
        \textbf{Total} & \textbf{6,878} & \textbf{190,778} & \textbf{4} & \textbf{5} & \textbf{12} & \textbf{8} \\
        \bottomrule
    \end{tabular}%
    }
    \vspace{-0.2cm}
\end{table}

% \begin{table*}[t]
%     \centering
%     \caption{\textbf{Composition of WorldUAV.} Video Clips and Train Images report the numbers of videos and selected training images in each background scene. Lighting, Weather, UAV Model, and Viewpoint report the numbers of distinct categories observed in the corresponding video metadata.}
%     \label{tab:worlduav_composition}
%     \small
%     \begin{tabular*}{\textwidth}{@{\extracolsep{\fill}}lrrrrrr@{}}
%         \toprule
%         Background Sence & Video Clips & Train Images & Lighting & Weather & UAV Model & Viewpoint \\
%         \midrule
%         Rural/pastoral & 1,833 & 58,119 & 4 & 5 & 4 & 7 \\
%         Forest/mountain & 1,440 & 36,290 & 4 & 5 & 8 & 7 \\
%         Coastal/water & 1,150 & 27,136 & 4 & 5 & 8 & 7 \\
%         Urban building & 922 & 25,390 & 4 & 5 & 8 & 7 \\
%         Urban road & 613 & 16,309 & 4 & 5 & 8 & 6 \\
%         Grassland/pasture & 465 & 15,915 & 3 & 4 & 4 & 5 \\
%         Park/recreation & 455 & 11,619 & 4 & 5 & 8 & 7 \\
%         \midrule
%         \textbf{Total} & \textbf{6,878} & \textbf{190,778} & \textbf{4} & \textbf{5} & \textbf{12} & \textbf{7} \\
%         \bottomrule
%     \end{tabular*}
% \end{table*}

\begin{table}[t]
    \centering
    \caption{\textbf{Spatiotemporal characteristics of WorldUAV.} This table summarizes the spatial and temporal properties of the dataset.}
    \vspace{-0.2cm}
    \label{tab:worlduav_specifications}
    \small
    \setlength{\tabcolsep}{4pt}
    \begin{tabular}{ll}
        \toprule
        Property & Characteristic \\
        \midrule
        Video frame rate & 25 fps \\
        Nominal clip length & 125 frames / 5 s \\
        Observed clip length & 10--125 frames / 0.4--5 s \\
        % Full-length clips & 6,342 / 6,878 (92.2\%) \\
        % Full-frame resolutions & 58 video / 53 image sizes \\
        Minimum resolution & $848\times480$ \\
        Maximum resolution & $4096\times2160$ \\
        Synthesis region & $832\times480$ pixels\\
        \bottomrule
    \end{tabular}
    \vspace{-0.3cm}
\end{table}

% WorldUAV statistics: all final bounding boxes in labels/{train,val_world,test_world}.
% Area ratio = normalized width * normalized height.
% Aspect ratio = (normalized width * image width) / (normalized height * image height).
% Image dimensions are read from the corresponding image files.
% Full-precision results and reproduction scripts: build/geometry_comparison/.
% DUT Anti-UAV and SynDroneVision values are transcribed from the user-provided table.
% Required packages: booktabs, graphicx.
\begin{table}[t]
    \centering
    \caption{\textbf{Comparison of object scale and aspect ratio.} Object scale is measured as bounding-box area divided by image area; aspect ratio is bounding-box width divided by height.}
    \vspace{-0.2cm}
    \label{tab:worlduav_geometry_comparison}
    \footnotesize
    \setlength{\tabcolsep}{4pt}
    \resizebox{0.9\columnwidth}{!}{%
    \begin{tabular}{lcccccc}
        \toprule
        & \multicolumn{3}{c}{Object Area Ratio} & \multicolumn{3}{c}{Object Aspect Ratio} \\
        \cmidrule(lr){2-4}\cmidrule(lr){5-7}
        Split & \multicolumn{1}{c}{min} & \multicolumn{1}{c}{mean} & \multicolumn{1}{c}{max} & \multicolumn{1}{c}{min} & \multicolumn{1}{c}{mean} & \multicolumn{1}{c}{max} \\
        \midrule
        \multicolumn{7}{l}{\textbf{DUT Anti-UAV}} \\
        Train & 0.000026 & 0.013 & 0.700 & 1.0 & 1.910 & 5.420 \\
        Val   & 0.000002 & 0.013 & 0.690 & 1.0 & 1.910 & 6.670 \\
        Test  & 0.000041 & 0.014 & 0.470 & 1.0 & 1.920 & 5.090 \\
        \midrule
        \multicolumn{7}{l}{\textbf{SynDroneVision}} \\
        Train & 0.001 & 0.322 & 1.0 & 0.021 & 1.291 & 9.993 \\
        Val   & 0.006 & 0.323 & 1.0 & 0.020 & 1.330 & 9.855 \\
        Test  & 0.009 & 0.323 & 1.0 & 0.041 & 1.302 & 8.383 \\
        \midrule
        \multicolumn{7}{l}{\textbf{WorldUAV}} \\
        Train & 0.000136 & 0.0056 & 0.122 & 1.235 & 1.852 & 3.678 \\
        Val & 0.000235 & 0.0088 & 0.101 & 1.235 & 1.872 & 2.590 \\
        Test & 0.000213 & 0.0086 & 0.092 & 1.239 & 1.868 & 2.814 \\
        \bottomrule
    \end{tabular}%
    }
    \vspace{-0.2cm}
\end{table}

% Source: all released video annotations in WorldUAV_Video/information.
% Eligible: exactly one box per frame and at least two frames; actual dimensions
% and complete frame coverage are verified against each video with ffprobe.
% Each video contributes once. Full-precision and per-video statistics:
% build/temporal_variation/statistics.json and per_video_metrics.csv.
% Reproduce: /usr/bin/python3 build/temporal_variation/compute_statistics.py
%           /usr/bin/python3 build/temporal_variation/make_table.py
\begin{table}[t]
    \centering
    \caption{\textbf{Temporal variation in WorldUAV.} Vertical and horizontal center spans are normalized by video height and width, respectively. The area ratio is the minimum bounding-box area divided by the maximum within a video.}
    \vspace{-0.2cm}
    \label{tab:worlduav_temporal_variation}
    \small
    \setlength{\tabcolsep}{4pt}
    \resizebox{\columnwidth}{!}{%
    \begin{tabular}{lcccc}
        \toprule
        Metric & min & mean & median & max \\
        \midrule
        Normalized vertical span ($S_H$) & 0.000 & 0.166 & 0.129 & 0.903 \\
        Normalized horizontal span ($S_W$) & 0.000 & 0.341 & 0.302 & 0.889 \\
        Min-to-max area ratio ($R_A$) & 0.013 & 0.578 & 0.598 & 1.000 \\
        \bottomrule
    \end{tabular}%
    }
    \vspace{-0.3cm}
\end{table}

\vspace{0.2cm}
\noindent
\textbf{UAV Model and Viewpoint.}
To ensure diversity in UAV appearance, size, and functionality, we select 12 representative models: \emph{Autel EVO II Pro V3}, \emph{DJI Agras T50}, \emph{DJI Air 3S}, \emph{DJI Avata 2}, \emph{DJI FlyCart 100}, \emph{DJI Matrice 350 RTK}, \emph{DJI Mavic 3 Pro}, \emph{DJI Mini 4 Pro}, \emph{Dreameagle X6-40 40L 6}, \emph{Parrot ANAFI USA}, \emph{Skydio X10}, and \emph{DJI Phantom 2}, as shown in Fig.~\ref{fig:visualization_uav}.
Previous UAV datasets have largely overlooked viewpoint due to the limited controllability of data construction. In contrast, our world model preserves the reference viewpoint while allowing physically plausible viewpoint variations during synthesis. We observe that the overall viewpoint category remains consistent across generated frames, enabling reliable viewpoint annotations. Specifically, as shown in Fig.~\ref{fig:visualization_viewpoint}, we define eight viewpoint categories: \emph{frontal-level (FL)}, \emph{frontal high-angle (FT)}, \emph{frontal low-angle (FB)}, \emph{side-level (SL)}, \emph{side high-angle (ST)}, \emph{side low-angle (SB)}, \emph{top-down (T)}, and \emph{bottom-up (B)}.
This facilitates UAV detection for specific viewpoints.

% Requires: \usepackage{booktabs,graphicx}
% Full-width, color-free table for a two-column CVPR/ICCV/ECCV-style paper.
% All numerical results reproduce the supplied table (three decimal places).
\begin{table*}[t]
    \centering
    \caption{\textbf{UAV detection performance across different training datasets.}
    We evaluate YOLOv8, YOLOv9, and YOLOv12 architectures under different training data configurations.
    @0.25, @0.50, and @0.50:0.95 denote mAP at the corresponding IoU thresholds.
    SDV denotes SynDroneVision, while SDV \& DUT and WorldUAV \& DUT indicate joint training on the corresponding datasets.}
    \vspace{-0.3cm}
    \label{tab:worlduav_yolo_detection}
    \footnotesize
    \setlength{\tabcolsep}{1.5pt}
    \renewcommand{\arraystretch}{0.90}
    \resizebox{\textwidth}{!}{%
    \begin{tabular}{@{}lccccc!{\hspace{3em}}lccccc@{}}
        \toprule
        Training data
        & $@0.25\uparrow$
        & $@0.50\uparrow$
        & $@0.50\!:\!0.95\uparrow$
        & $\mathrm{FNR}\downarrow$
        & $\mathrm{FDR}\downarrow$
        & Training data
        & $@0.25\uparrow$
        & $@0.50\uparrow$
        & $@0.50\!:\!0.95\uparrow$
        & $\mathrm{FNR}\downarrow$
        & $\mathrm{FDR}\downarrow$ \\
        \midrule
        \multicolumn{6}{c}{\textbf{YOLOv8m}} & \multicolumn{6}{c}{\textbf{YOLOv8l}} \\
        \cmidrule(l{1pt}r{25pt}){1-6}\cmidrule(l{1pt}r{1pt}){7-12}
        DUT Anti-UAV           & 0.956 & 0.933 & 0.669 & 0.118 & 0.021 & DUT Anti-UAV           & 0.922 & 0.896 & 0.628 & 0.149 & 0.067 \\
        SynDroneVision         & 0.677 & 0.639 & 0.422 & 0.525 & 0.103 & SynDroneVision         & 0.746 & 0.716 & 0.468 & 0.438 & 0.079 \\
        SDV \& DUT             & 0.960 & 0.938 & 0.686 & 0.117 & 0.025 & SDV \& DUT             & 0.963 & 0.944 & 0.696 & 0.105 & \textbf{0.015} \\
        WorldUAV               & 0.781 & 0.758 & 0.520 & 0.364 & 0.028 & WorldUAV               & 0.792 & 0.768 & 0.520 & 0.357 & 0.030 \\
        \textbf{WorldUAV \& DUT} & \textbf{0.974} & \textbf{0.953} & \textbf{0.694} & \textbf{0.079} & \textbf{0.017} & \textbf{WorldUAV \& DUT} & \textbf{0.979} & \textbf{0.957} & \textbf{0.702} & \textbf{0.074} & 0.016 \\
        \addlinespace[3pt]
        \multicolumn{6}{c}{\textbf{YOLOv9c}} & \multicolumn{6}{c}{\textbf{YOLOv9e}} \\
        \cmidrule(l{1pt}r{25pt}){1-6}\cmidrule(l{1pt}r{1pt}){7-12}
        DUT Anti-UAV           & 0.959 & 0.935 & 0.668 & 0.123 & 0.023 & DUT Anti-UAV           & 0.944 & 0.915 & 0.643 & 0.149 & 0.042 \\
        SynDroneVision         & 0.700 & 0.666 & 0.429 & 0.474 & 0.093 & SynDroneVision         & 0.767 & 0.733 & 0.460 & 0.405 & 0.055 \\
        SDV \& DUT             & 0.961 & 0.941 & \textbf{0.695} & 0.107 & 0.022 & SDV \& DUT & 0.969 & 0.955 & 0.723 & 0.095 & \textbf{0.009} \\
        WorldUAV               & 0.810 & 0.782 & 0.531 & 0.340 & 0.022 & WorldUAV               & 0.839 & 0.810 & 0.559 & 0.310 & 0.029 \\
        \textbf{WorldUAV \& DUT} & \textbf{0.977} & \textbf{0.955} & \textbf{0.695} & \textbf{0.089} & \textbf{0.018} & \textbf{WorldUAV \& DUT} & \textbf{0.988} & \textbf{0.966} & \textbf{0.725} & \textbf{0.066} & 0.019 \\
        \addlinespace[3pt]
        \multicolumn{6}{c}{\textbf{YOLOv12l}} & \multicolumn{6}{c}{\textbf{YOLOv12x}} \\
        \cmidrule(l{1pt}r{25pt}){1-6}\cmidrule(l{1pt}r{1pt}){7-12}
        DUT Anti-UAV           & 0.972 & 0.937 & 0.655 & 0.097 & 0.043 & DUT Anti-UAV           & 0.978 & 0.946 & 0.654 & 0.079 & 0.046 \\
        SynDroneVision         & 0.767 & 0.736 & 0.450 & 0.562 & 0.169 & SynDroneVision         & 0.804 & 0.768 & 0.461 & 0.456 & 0.173 \\
        SDV \& DUT             & 0.973 & 0.951 & 0.675 & 0.084 & 0.040 & SDV \& DUT             & 0.981 & 0.962 & 0.683 & 0.075 & 0.020 \\
        WorldUAV               & 0.817 & 0.784 & 0.532 & 0.318 & 0.031 & WorldUAV               & 0.819 & 0.788 & 0.534 & 0.322 & 0.027 \\
        \textbf{WorldUAV \& DUT} & \textbf{0.979} & \textbf{0.957} & \textbf{0.700} & \textbf{0.073} & \textbf{0.023} & \textbf{WorldUAV \& DUT} & \textbf{0.987} & \textbf{0.963} & \textbf{0.707} & \textbf{0.068} & \textbf{0.018} \\
        \bottomrule
    \end{tabular}%
    }
    \vspace{-0.2cm}
\end{table*}

\begin{figure*}[t]
    \centering
    \includegraphics[width=\textwidth]{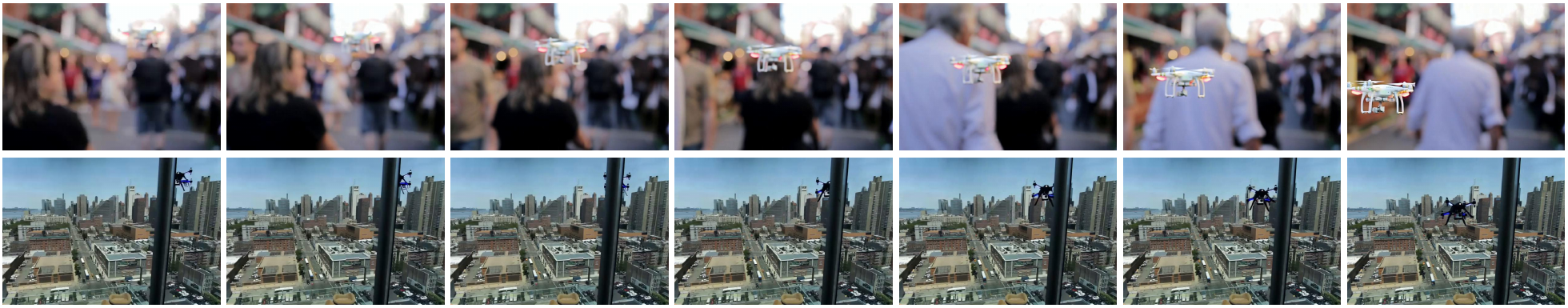}
    \vspace{-0.5cm}
    \caption{\textbf{Scene-consistent UAV synthesis.}
    Each row shows temporally ordered frames from a WorldUAV video.
    Top: depth-dependent sharpness, with the synthesized UAV becoming clearer as it approaches the camera.
    Bottom: foreground occlusion, with the UAV partially hidden behind a pillar as it traverses the scene.}
    \label{fig:worlduav_physical_consistency}
    \vspace{-0.4cm}
\end{figure*}

\subsection{Dataset Composition}
\label{sec:dataset_composition}

For the construction of WorldUAV, UAV models and viewpoints are selected in relation to the background scene, while lighting and weather reflect the conditions of the source videos.
For rural/pastoral and grassland/pasture scenes, we restrict the candidate models to DJI Matrice 350 RTK, DJI Agras T50, Dreameagle X6-40, and DJI FlyCart 100, reflecting the suitability of these platforms for agricultural and field operations.
The remaining five scene categories use the other eight models.
Within these scene-dependent candidate sets, the MLLM interprets the background semantics, spatial layout, and camera perspective to select the UAV model and viewpoint most appropriate for the environment.
This selection strategy combines appearance diversity with plausible associations.

WorldUAV comprises 6,878 videos across seven background scenes, with per-scene counts ranging from 455 for park/recreation to 1,833 for rural/pastoral, as shown in Tab.~\ref{tab:worlduav_composition}.
From these synthesized videos, we construct an image dataset of 205,778 images, partitioned into 190,778 training, 5,000 validation, and 10,000 test images.
Each selected image retains its parent video identifier and frame index.
The images follow a common annotation schema comprising UAV bounding boxes, resolution, background scene, lighting, weather, and the synthesis prompt, together with UAV model and viewpoint labels.
Overall, WorldUAV provides the largest image collection among publicly available UAV detection datasets and the largest video collection among tracking datasets compared in Tab.~\ref{tab:dataset_comparison}.

\subsection{Dataset Characteristic}
\label{sec:dataset_characteristics}

We summarize the key characteristics of WorldUAV dataset as follows:

\noindent
\textbf{Spatiotemporal Characteristics.}
As shown in Tab.~\ref{tab:worlduav_specifications}, all videos are encoded at 25 fps.
Video lengths range from 10 to 124 frames.
% The collection contains 58 distinct full-frame resolutions, while the selected images cover 53.
% The most common image resolutions are $1920\times1080$, $3840\times2160$, $854\times480$, and $1280\times720$, accounting for 26.1\%, 21.7\%, 18.9\%, and 10.6\% of images, respectively. 
Image resolutions range from $848\times480$ to $4096\times2160$.
The world model performs editing and synthesis within a fixed spatial region of $832\times480$ pixels.

\begin{table*}[t]
    \centering
    \caption{\textbf{Out-of-distribution UAV detection performance.} We evaluate YOLOv12l detectors trained on different datasets using two out-of-distribution benchmarks, UAV-Eagle~\cite{barivsic2022brain} and Drone Dataset~\cite{aksoy2019drone}.}
    \vspace{-0.3cm}
    \label{tab:worlduav_generalization}
    \footnotesize
    \setlength{\tabcolsep}{1.5pt}
    \renewcommand{\arraystretch}{0.90}
    \resizebox{\textwidth}{!}{%
    \begin{tabular}{@{}lcccccc!{\hspace{1.5em}}lcccccc@{}}
        \toprule
        \multicolumn{7}{c}{\textbf{UAV-Eagle}} & \multicolumn{7}{c}{\textbf{Drone Dataset}} \\
        \cmidrule(l{1pt}r{13pt}){1-7}\cmidrule(l{1pt}r{1pt}){8-14}
        Training data
        & $\mathrm{Precision}\uparrow$
        & $\mathrm{Recall}\uparrow$
        & $@0.50\uparrow$
        & $@0.50\!:\!0.95\uparrow$
        & $\mathrm{FNR}\downarrow$
        & $\mathrm{FDR}\downarrow$
        & Training data
        & $\mathrm{Precision}\uparrow$
        & $\mathrm{Recall}\uparrow$
        & $@0.50\uparrow$
        & $@0.50\!:\!0.95\uparrow$
        & $\mathrm{FNR}\downarrow$
        & $\mathrm{FDR}\downarrow$ \\
        \cmidrule(l{1pt}r{13pt}){1-7}\cmidrule(l{1pt}r{1pt}){8-14}
        DUT Anti-UAV             & 0.873 & 0.745 & 0.773 & 0.273 & 0.238 & 0.203 & DUT Anti-UAV             & 0.571 & 0.558 & 0.493 & 0.171 & 0.398 & 0.529 \\
        SynDroneVision           & 0.801 & 0.779 & 0.785 & 0.299 & 0.286 & 0.290 & SynDroneVision           & 0.531 & 0.558 & 0.428 & 0.131 & 0.413 & 0.574 \\
        SDV \& DUT               & 0.858 & \textbf{0.794} & 0.812 & 0.278 & 0.252 & 0.157 & SDV \& DUT               & 0.580 & 0.575 & 0.505 & 0.172 & 0.369 & 0.494 \\
        WorldUAV                 & 0.863 & 0.780 & 0.795 & 0.304 & 0.206 & 0.165 & WorldUAV                 & 0.583 & 0.548 & 0.468 & 0.146 & 0.422 & 0.479 \\
        {WorldUAV \& DUT} & \textbf{0.879} & 0.790 & \textbf{0.837} & \textbf{0.313} & \textbf{0.203} & \textbf{0.131} & {WorldUAV \& DUT} & \textbf{0.586} & \textbf{0.612} & \textbf{0.523} & \textbf{0.182} & \textbf{0.363} & \textbf{0.457} \\
        \bottomrule
    \end{tabular}%
    }
    \vspace{-0.1cm}
\end{table*}

\begin{figure*}[t]
    \centering
    \includegraphics[width=\textwidth]{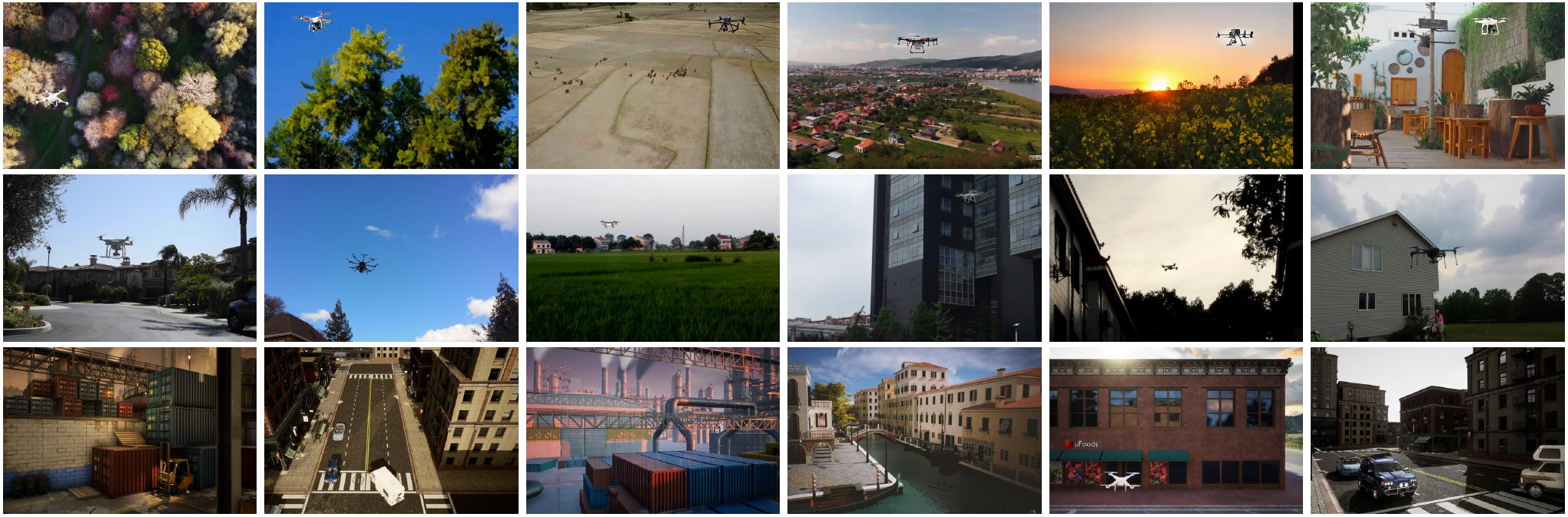}
    \vspace{-0.6cm}
    \caption{\textbf{Visual comparison of synthetic and real UAV images.}
    Top: WorldUAV; middle: DUT Anti-UAV~\cite{zhao2022vision}; bottom: SynDroneVision~\cite{lenhard2025syndronevision}.
    The examples illustrate the closer photographic appearance of WorldUAV to real observations.}
    \label{fig:worlduav_synthesis_comparison}
    \vspace{-0.3cm}
\end{figure*}

\noindent
\textbf{Object Scale.}
WorldUAV preserves full-frame scene context, with mean bounding-box-to-image area ratios of 0.0056--0.0088 across splits (Tab.~\ref{tab:worlduav_geometry_comparison}).
These means are closer to DUT Anti-UAV~\cite{zhao2022vision} (0.013--0.014) than SynDroneVision~\cite{lenhard2025syndronevision} (0.322--0.323), indicating better alignment with real-world data in mean relative object scale.

\noindent
\textbf{Aspect Ratio.}
WorldUAV has mean bounding-box width-to-height ratios of 1.852--1.872, close to DUT Anti-UAV (1.910--1.920).
Compared with SynDroneVision, it exhibits less extreme proportions and closer agreement with DUT in mean aspect ratio (Tab.~\ref{tab:worlduav_geometry_comparison}).

\noindent
\textbf{Temporal Variation.}
We measure horizontal and vertical bounding-box-center spans normalized by video width and height, respectively, and the minimum-to-maximum bounding-box area ratio.
Smaller area ratios indicate greater scale variation.
As shown in Tab.~\ref{tab:worlduav_temporal_variation}, vertical and horizontal spans reach 90.3\% and 88.9\%, while the area ratio reaches a minimum of 0.013, demonstrating substantial variation in UAV position and apparent scale.

% Required in the main manuscript preamble:
% \usepackage{booktabs}
% \usepackage{multirow}
%
% Single-column, color-free CVPR/ICCV/ECCV-style table.
% The tabular* environment fixes the width to \columnwidth;
% change \footnotesize to \scriptsize or \small as needed.
% Unlike \resizebox, tabular* does not scale text to fill the width.
\begin{table}[t]
    \centering
    \caption{Precision and recall of YOLOv12 detectors. }
    \vspace{-0.3cm}
    \label{tab:worlduav_yolov12_pr}
    \footnotesize
    \setlength{\tabcolsep}{3pt}
    \renewcommand{\arraystretch}{1.08}
    \begin{tabular*}{0.95\columnwidth}{@{\extracolsep{\fill}}llcc@{}}
        \toprule
        YOLO Model & Training Data & Precision $\uparrow$ & Recall $\uparrow$ \\
        \midrule
        \multirow{5}{*}{\textbf{YOLOv12l}}
          & DUT Anti-UAV   & 0.957 & 0.901 \\
          & SynDroneVision & 0.870 & 0.695 \\
          & SDV \& DUT      & 0.959 & 0.928 \\
          & WorldUAV       & 0.947 & 0.707 \\
          & \textbf{WorldUAV \& DUT} & \textbf{0.969} & \textbf{0.936} \\
        \midrule
        \multirow{5}{*}{\textbf{YOLOv12x}}
          & DUT Anti-UAV   & 0.960 & 0.915 \\
          & SynDroneVision & 0.838 & 0.739 \\
          & SDV \& DUT      & 0.965 & 0.936 \\
          & WorldUAV       & 0.943 & 0.715 \\
          & \textbf{WorldUAV \& DUT} & \textbf{0.971} & \textbf{0.945} \\
        \bottomrule
    \end{tabular*}
    \vspace{-0.3cm}
\end{table}

\section{Experiments}
\label{sec:experiments}

\subsection{Experimental Setup}
\label{sec:experimental_setup}
We compare WorldUAV with real-world DUT Anti-UAV (DUT)~\cite{zhao2022vision} and synthetic SynDroneVision (SDV)~\cite{lenhard2025syndronevision} using YOLOv8m/l, YOLOv9c/e, and YOLOv12l/x.
Detection is evaluated on the DUT test set.
An additional comparison matches the synthetic training set sizes to assess data quality while controlling for data volume.
Following SDV, each model is trained for 100 epochs with a batch size of 64, except for YOLOv9e, which uses a batch size of 32. All other hyperparameters follow the default configurations specified in~\cite{ultralytics2024yolov8}.
For tracking evaluation on UAVSwarmV2~\cite{wang2026topology}, we replace the detectors in TrackTrack~\cite{shim2025focusing} and BoostTrack++~\cite{stanojevic2025boosttrack++} with YOLOv12l/x trained separately on each dataset.

Cross-dataset generalization is evaluated on UAV-Eagle~\cite{barivsic2022brain} and Drone Dataset~\cite{aksoy2019drone}.
Since UAV-Eagle has been updated since the SDV evaluation, the published YOLOv9e results are not directly comparable.
We therefore reproduce the SDV baselines using the more recent YOLOv12l and evaluate all configurations on the current benchmarks.

Detection metrics comprise precision, recall, mAP at IoU thresholds of 0.25 and 0.50 and averaged over 0.50--0.95, false negative rate (FNR), and false discovery rate (FDR).
Tracking metrics comprise higher-order tracking accuracy (HOTA), detection accuracy (DetA), identity F1 score (IDF1), and the number of identity switches (IDSW).

\subsection{Detection and Tracking}

\paragraph{UAV Detection.}
\label{sec:uav_detection}
We evaluate six YOLO variants on the DUT Anti-UAV test set under different training configurations, as shown in Tab.~\ref{tab:worlduav_yolo_detection}.
Combining WorldUAV with DUT consistently improves mAP and reduces FNR and FDR over DUT-only training.
For example, YOLOv8l improves from 89.6\% to 95.7\% mAP at IoU 0.5.
When used alone for training, WorldUAV outperforms SDV in mAP across all six detectors; for YOLOv8m, mAP at IoU 0.5 increases from 63.9\% to 75.8\%.
Nevertheless, its mAP remains lower than that of DUT-only training due to differences in data distributions.
Tab.~\ref{tab:worlduav_yolov12_pr} further shows the advantages of WorldUAV in terms of precision and recall.

% Requires in the preamble: \usepackage{booktabs,multirow}
% Single-column table for CVPR/ICCV/ECCV-style two-column manuscripts.
% Metrics: one decimal place except integer IDSW.
% Best results are bolded within each detector-model/tracker combination
% across the three training datasets (higher is better, except IDSW).
% Uses \columnwidth directly without \resizebox or \scalebox.
\begin{table}[t]
    \centering
    \caption{\textbf{UAV tracking performance with detectors trained on different datasets.}
    The detectors in TrackTrack and BoostTrack++ are replaced. }
    \vspace{-0.3cm}
    \label{tab:worlduav_tracking_comparison}
    \footnotesize
    \setlength{\tabcolsep}{1.4pt}
    \renewcommand{\arraystretch}{1.12}
    \begin{tabular*}{\columnwidth}{@{\extracolsep{\fill}}llccccc@{}}
        \toprule
        \multicolumn{2}{c}{Detector}
          & \multirow{2}{*}{Tracker}
          & \multirow{2}{*}{HOTA$\uparrow$}
          & \multirow{2}{*}{DetA$\uparrow$}
          & \multirow{2}{*}{IDF1$\uparrow$}
          & \multirow{2}{*}{IDSW$\downarrow$} \\
        \cmidrule(lr){1-2}
        Model & Dataset & & & & & \\
        \midrule
        \multirow{6}{*}{\bf YOLOv12l}
          & DUT            & \multirow{3}{*}{TrackTrack}   & 33.9 & 25.2 & 31.7 & 166 \\
          & SDV            &                              & 34.2 & 25.6 & 29.2 & 176 \\
          & Ours       &                              & \textbf{34.5} & \textbf{25.8} & \textbf{35.3} & \textbf{130} \\
        \cmidrule(lr){2-7}
          & DUT            & \multirow{3}{*}{BoostTrack++} & 34.3 & 25.6 & 32.5 & 155 \\
          & SDV            &                              & 34.4 & \textbf{25.8} & 30.3 & 162 \\
          & Ours       &                              & \textbf{34.8} & 25.7 & \textbf{36.1} & \textbf{111} \\
        \midrule
        \multirow{6}{*}{\bf YOLOv12x}
          & DUT            & \multirow{3}{*}{TrackTrack}   & 33.9 & 26.5 & 30.6 & 189 \\
          & SDV            &                              & 34.2 & 26.1 & 29.7 & 176 \\
          & Ours       &                              & \textbf{34.5} & \textbf{27.0} & \textbf{35.3} & \textbf{134} \\
        \cmidrule(lr){2-7}
          & DUT            & \multirow{3}{*}{BoostTrack++} & 33.7 & 25.1 & 32.3 & 148 \\
          & SDV            &                              & 33.7 & 25.3 & 30.6 & 144 \\
          & Ours       &                              & \textbf{34.0} & \textbf{25.6} & \textbf{35.3} & \textbf{125} \\
        \bottomrule
    \end{tabular*}
    \vspace{-0.3cm}
\end{table}

\paragraph{UAV Tracking.}
\label{sec:uav_tracking}
We evaluate TrackTrack and BoostTrack++ using YOLOv12l and YOLOv12x detectors trained separately on DUT, SDV, and WorldUAV, as shown in Tab.~\ref{tab:worlduav_tracking_comparison}.
Across all detector--tracker combinations, WorldUAV achieves the highest HOTA and IDF1 and the fewest identity switches relative to DUT and SDV.
For example, with YOLOv12l and BoostTrack++, identity switches decrease from 155 with DUT to 111 with WorldUAV.
These results show that detectors trained on WorldUAV improve downstream tracking accuracy and identity consistency.

\subsection{Analysis}

\vspace{0.2cm}
\noindent
\textbf{Cross-Dataset Generalization.}
\label{sec:cross_dataset_generalization}
As shown in Tab.~\ref{tab:worlduav_generalization}, training solely on WorldUAV yields higher mAP than training on SDV.
Moreover, combining WorldUAV with DUT achieves the highest mAP and the lowest FNR and FDR.
These results indicate that WorldUAV complements real training data and improves detection robustness across datasets.

\vspace{0.2cm}
\noindent
\textbf{Construction Efficiency.}
\label{sec:construction_efficiency}
Tab.~\ref{tab:worlduav_construction_cost} reports the processing time and API cost of constructing a UAV video with WorldUAV Agent, amortized over its frames.
Source acquisition and video compositing run locally, with world-model inference performed on an NVIDIA H100 GPU with 80\,GB of memory.
Monetary costs include only MLLM API calls for planning and verification, excluding local computation costs.
The results demonstrate efficient automated UAV data construction with low API expenditure.

\vspace{0.2cm}
\noindent
\textbf{Synthesis Visualization.} \label{sec:synthesis_visualization}
We visualize synthesized UAV images to examine their consistency with real scenes.
By combining real video backgrounds with the world model's learned physical and spatiotemporal priors, WorldUAV adapts UAV appearance to the surrounding scene and its imaging conditions.
Fig.~\ref{fig:worlduav_physical_consistency} illustrates two manifestations of this capability: depth-dependent sharpness and foreground occlusion.
In the top sequence, the UAV becomes progressively sharper as it approaches the camera, consistent with the scene's depth-of-field characteristics.
In the bottom sequence, the UAV is partially occluded as it passes behind a foreground pillar, respecting the relative depth ordering of scene elements.
These examples demonstrate the scene consistency of WorldUAV.

Fig.~\ref{fig:worlduav_synthesis_comparison} compares WorldUAV with real images from DUT Anti-UAV~\cite{zhao2022vision} and synthetic images from SDV~\cite{lenhard2025syndronevision}.
Our examples resemble real UAV observations more closely in overall photographic appearance, whereas the game-engine-rendered SDV examples exhibit more pronounced differences in scene texture and lighting.

\begin{table}[t]
    \centering
    \caption{\textbf{Construction efficiency of WorldUAV.}
    Per-image time and reported API cost for the four construction stages.
    Video-level processing is amortized over images.}
    \vspace{-0.2cm}
    \label{tab:worlduav_construction_cost}
    \small
    \setlength{\tabcolsep}{3pt}
    \renewcommand{\arraystretch}{1.12}
    \begin{tabular*}{\columnwidth}{@{\extracolsep{\fill}}lrr@{}}
        \toprule
        Pipeline stage & Time (s/image) & API cost (USD/image) \\
        \midrule
        Source preparation    & 0.70  & \textemdash \\
        Scene-aware planning  & 1.44  & 0.00025 \\
        World-model synthesis & 11.90 & \textemdash \\
        Quality verification  & 0.96  & 0.00025 \\
        \midrule
        \textbf{Total} & \textbf{15.00} & \textbf{0.00050} \\
        \bottomrule
    \end{tabular*}
    \vspace{-0.2cm}
\end{table}

% Requires in the main document preamble:
% \usepackage{booktabs,graphicx,multirow}
% Single-column, color-free table for CVPR/ICCV/ECCV-style papers.
% All reported metrics are rounded to three decimal places.
% Best results within each detector are highlighted in bold.
\begin{table}[t]
    \centering
    \caption{\textbf{Dataset quality evaluation with equal synthetic training set sizes.}
    Detectors are jointly trained on the same DUT training set and either SDV or a WorldUAV subset containing an equal number of synthetic images.}
    \vspace{-0.2cm}
    \label{tab:worlduav_joint_training_detection}
    \footnotesize
    \setlength{\tabcolsep}{2pt}
    \renewcommand{\arraystretch}{0.95}
    \resizebox{\columnwidth}{!}{%
    \begin{tabular}{@{}llccccc@{}}
        \toprule
        Model & Training data
        & $@0.25\uparrow$
        & $@0.50\uparrow$
        & $@0.50\!:\!0.95\uparrow$
        & $\mathrm{FNR}\downarrow$
        & $\mathrm{FDR}\downarrow$ \\
        \midrule
        \multirow{2}{*}{{YOLOv8l}}
          & SDV \& DUT               & 0.963 & 0.944 & 0.696 & 0.105 & \textbf{0.015} \\
          & {WorldUAV \& DUT} & \textbf{0.974} & \textbf{0.950} & \textbf{0.701} & \textbf{0.089} & 0.016 \\
        \midrule
        \multirow{2}{*}{{YOLOv9e}}
          & SDV \& DUT               & 0.969 & 0.955 & 0.723 & 0.095 & \textbf{0.009} \\
          & {WorldUAV \& DUT} & \textbf{0.980} & \textbf{0.963} & \textbf{0.724} & \textbf{0.079} & 0.019 \\
        \bottomrule
    \end{tabular}%
    }
    \vspace{-0.3cm}
\end{table}

\vspace{0.2cm}
\noindent
\textbf{Qualitative Evaluation.}
\label{sec:qualitative_evaluation}
To assess dataset quality while controlling for data volume, we select a WorldUAV subset containing the same number of training images as SDV.
As shown in Tab.~\ref{tab:worlduav_joint_training_detection}, WorldUAV yields higher mAP for both detectors, supporting its greater utility as training data at the same synthetic data scale.

\section{Conclusion}
\label{sec:conclusion}
% We introduced WorldUAV, a dataset built by integrating UAVs into real-world videos using world models for realistic, temporally consistent synthesis.
% An MLLM-orchestrated agentic pipeline enables efficient, scalable, and customizable construction.
% Experiments demonstrate a reduced synthetic-to-real gap and improved UAV detection, tracking, and cross-dataset generalization.

We introduced WorldUAV, a dataset built by integrating UAVs into real-world videos using world models.
Their learned spatiotemporal priors enable physically plausible, temporally consistent synthesis while preserving scene realism.
WorldUAV Agent uses an MLLM to orchestrate source acquisition, trajectory planning, world-model invocation, and quality verification, enabling efficient, scalable, and customizable data construction.
Experiments demonstrate a reduced synthetic-to-real gap and improved UAV detection, tracking, and cross-dataset generalization.

% \newpage
% \section*{LLM Usage Statement}
% No LLMs were used in manuscript writing or experimental work. LLMs were used to assist with preliminary literature research, and all relevant information was independently verified by the authors.
{
    \small
    \bibliographystyle{ieeenat_fullname}
    \bibliography{main}
}

\clearpage
\setcounter{page}{1}
\maketitlesupplementary
\appendix

\section{Details on UAV Datasets}
\label{sec:appendix_detection_datasets}

Complementing Sec.~\ref{sec:related_uav_datasets} and Tab.~\ref{tab:dataset_comparison}, this appendix describes datasets for UAV detection, aerial-object recognition, and tracking.
We summarize their acquisition settings, composition, target diversity, and annotation protocols, covering both real-world recordings and synthetic imagery.
Alongside the detection resources in the comparison table, we include Anti-UAV and UAVSwarmV2 to characterize the data available for single-UAV and swarm tracking, respectively.

\vspace{0.2cm}
\noindent\textbf{USC Drone Detection and Tracking.}
The USC Drone Detection and Tracking dataset~\cite{chen2017deep,wang2019towards} contains 30 videos recorded on the University of Southern California campus using a DJI Phantom.
Each video lasts approximately one minute and is recorded at 15 frames per second (FPS) with a resolution of $1920\times1080$, yielding approximately 27,000 annotated frames.
The recordings cover changes in background, viewing angle, illumination, and target appearance, including fast motion and partial occlusion.
Although the sequences capture varied observation conditions, their UAV appearance diversity is limited to one model.

\vspace{0.2cm}
\noindent\textbf{Drone Dataset.}
The Drone Dataset released by Aksoy et al.~\cite{aksoy2019drone} comprises 4,012 annotated images, with 3,611 images for training and 401 for testing.
The images are collected from online videos and image searches and span resolutions from $300\times168$ to $3840\times2160$.
DJI Phantom drones are depicted across heterogeneous backgrounds and viewpoints.
The collection provides an image-based resource for UAV detection; its use of video frames as an acquisition source does not establish a temporally continuous tracking benchmark.

\vspace{0.2cm}
\noindent\textbf{MAV-VID.}
The Multirotor Aerial Vehicle VID dataset~\cite{rodriguez2020adaptive} contains 64 videos and 40,232 annotated frames, divided into 29,500 training and 10,732 validation images.
The sequences depict individual drones and are acquired using handheld devices, ground-based cameras, and UAV-mounted cameras.
This mixture of recording platforms introduces variation in camera motion, target scale, and background composition.
The average UAV bounding-box size is approximately $136\times77$ pixels, making the dataset relevant to detecting relatively small aerial targets in video.

\vspace{0.2cm}
\noindent\textbf{Det-Fly.}
Det-Fly~\cite{zheng2021air} targets air-to-air detection and contains 13,271 images at a resolution of $3840\times2160$.
A camera carried by another UAV records a DJI Mavic against sky, urban, field, and mountain backgrounds.
Images are sampled from videos at 5 FPS or acquired at selected relative positions, covering frontal, overhead, and upward-looking views.
Nearly half of the target instances occupy less than 5\% of the image area.
The dataset emphasizes scale and viewpoint variation while retaining a single target model.

\vspace{0.2cm}
\noindent\textbf{UAV-Eagle.}
UAV-Eagle~\cite{barivsic2022brain} comprises 510 annotated images collected with a UAV-mounted camera for evaluating aerial target detection.
The images include one or multiple Eagle quadcopters against cluttered backgrounds containing vegetation, buildings, clouds, vehicles, and people.
Changes in viewpoint, lighting, and motion artifacts increase the difficulty of localizing the target.
Its small image collection provides a focused test of detection under challenging flight conditions, although the target appearance is restricted to the Eagle model.

\vspace{0.2cm}
\noindent\textbf{UAVData.}
UAVData~\cite{zeng2021uavdata} contains 13,803 annotated RGB images at $1280\times720$ resolution, covering six UAV models and balloon distractors.
Its composition includes 7,320 single-UAV images, 4,346 multiple-UAV images, and 2,137 balloon images.
Indoor recordings cover workshops and laboratories, while outdoor images include sky, vegetation, and buildings.
The dataset incorporates abrupt lighting changes and motion blur, with target sizes ranging from $5\times23$ to $720\times303$ pixels.
These properties support evaluation across substantial variations in target scale and scene complexity.

\vspace{0.2cm}
\noindent\textbf{Halmstad Data.}
The multi-sensor dataset of Svanstr\"om et al.~\cite{svanstrom2021dataset}, listed as Halmstadt Data in Tab.~\ref{tab:dataset_comparison}, includes 365 infrared and 285 visible-light videos, together with audio recordings.
The ten-second clips are primarily recorded in daylight at Swedish airport sites.
Across both visual modalities, the collection provides 203,328 annotated frames with drone, bird, airplane, and helicopter categories.
It includes several UAV models and aerial distractors, supporting the study of small-target recognition in surveillance settings.
The frame and video totals cover all annotated categories, rather than UAVs alone.

\vspace{0.2cm}
\noindent\textbf{DUT Anti-UAV.}
DUT Anti-UAV~\cite{zhao2022vision} provides separate detection and tracking subsets.
The detection subset contains 10,000 images, partitioned into 5,200 training, 2,600 validation, and 2,200 test images, with resolutions ranging from $240\times160$ to $5616\times3744$.
The collection covers 35 UAV types against backgrounds such as buildings, farmland, forests, and open sky.
It includes daytime, nighttime, dawn, and dusk observations, together with sunny, cloudy, and snowy conditions.
Small targets are prominent: the mean bounding-box area is approximately 1.3\% of the image area.
The tracking subset contains 20 videos.

\vspace{0.2cm}
\noindent\textbf{VisioDECT.}
VisioDECT~\cite{sakib2025visiodect} comprises 20,924 annotated RGB images at $852\times480$ resolution.
The collection covers six UAV models, including Anafi Extended, DJI FPV, DJI Phantom, EFT-E410S, Mavic2-Air, and Mavic2-Enterprise Zoom.
Recordings span different locations, altitudes, and capture times, organized around sunny, cloudy, and evening conditions.
Images without UAVs are removed during data cleaning, and the retained images are manually annotated.
The dataset therefore emphasizes detection across environmental and appearance variations within UAV-containing images.

\vspace{0.2cm}
\noindent\textbf{Malicious Drones.}
The dataset used by Jamil et al.~\cite{jamil2022distinguishing} contains 776 images standardized to $224\times224$ pixels, with 543 training and 233 test images.
The five categories are airplane, bird, drone, helicopter, and malicious drone, where the last category includes UAVs carrying potentially hazardous payloads.
The images exhibit low illumination, occlusion, reduced target visibility, and adverse weather.
This resource is designed for image classification; its inclusion in the comparison concerns aerial-object recognition and should not be interpreted as a bounding-box detection benchmark.

\vspace{0.2cm}
\noindent\textbf{S-UAV-T.}
S-UAV-T, introduced with Sim2Air~\cite{barisic2022sim2air}, is a synthetic dataset for UAV-to-UAV detection containing 52,500 rendered images.
Blender and its Cycles renderer are used to vary UAV models, object counts, scale, camera pose, and illumination.
Texture randomization assigns diverse, including unconventional, surface appearances to the UAVs, encouraging detectors to use shape cues.
The dataset covers daylight, partly cloudy, and twilight conditions, with the $608\times608$ image resolution reported in Tab.~\ref{tab:dataset_comparison}.
Its principal contribution is controlled appearance variation for transferring detection models from simulation to flight imagery.

\vspace{0.2cm}
\noindent\textbf{Drone-vs-Bird Detection Challenge.}
The Drone-vs-Bird benchmark~\cite{coluccia2024drone} provides 77 training videos and 30 additional test videos captured with static and moving cameras.
The recordings cover eight commercial UAV types, including fixed-wing and rotary-wing platforms, in urban, woodland, agricultural, and maritime environments.
Image resolutions range from $720\times576$ to $3840\times2160$.
Birds and other aerial objects introduce distractors, while small target extent, sun glare, and changes in weather and camera characteristics complicate detection.
Drone instances are annotated, whereas birds serve as unannotated distractors; test annotations are withheld.

\vspace{0.2cm}
\noindent\textbf{Anti-UAV.}
Anti-UAV~\cite{jiang2021anti} contains 318 paired RGB and thermal-infrared video sequences for single-UAV tracking.
The recordings include multiple commercial UAV models under daytime and nighttime illumination, with backgrounds such as buildings, trees, and clouds.
Frame-level bounding boxes support temporal localization across changes in target visibility and scale.
This dataset complements the image-based resources above through paired visual modalities and continuous observations, although its original protocol is designed for tracking rather than a separate image-level detection split.

\vspace{0.2cm}
\noindent\textbf{SynDroneVision.}
SynDroneVision~\cite{lenhard2025syndronevision} uses Unreal Engine to render UAVs in eight virtual environments, including urban, industrial, agricultural, and recreational scenes.
It comprises 72 annotated image sequences with 13 UAV models and varied camera positions, backgrounds, and illumination.
Realistic model textures and controlled lighting are complemented by blurred image variants to diversify appearance.
The reported splits contain 131,238 training, 8,800 validation, and 4,000 test images, with bounding boxes provided in YOLO format.
The dataset offers systematic control over scene composition and serves as the simulation-based comparison in our experiments.
% The supplied PDF and comparison table report 140,038 total images, but the
% stated splits sum to 144,038. Retain the source splits without repeating the
% inconsistent total; resolving the source table is outside this appendix edit.

\vspace{0.2cm}
\noindent\textbf{RealDroneVision.}
RealDroneVision~\cite{sivapuram2026realdronevision} contains 173,023 real-world UAV images, with 138,417 training and 34,606 test images at resolutions up to $1920\times1080$.
Its annotation pipeline combines tracker-generated proposals, human filtering, detector-based recovery of missed instances, and iterative manual refinement.
The imagery covers indoor and outdoor scenes, varied UAV appearances, and distractors such as birds and kites.
The mean target area is approximately 0.1\% of the image area, emphasizing small-object detection.
The images originate from videos, while the reported evaluation protocol operates on detection images.

\vspace{0.2cm}
\noindent\textbf{UAVSwarmV2.}
UAVSwarmV2~\cite{wang2026topology} extends UAVSwarm for detection and multi-object tracking of low-altitude UAV swarms.
The extension adds 24 sequences containing 8,841 frames captured at 30 FPS.
The resulting tracking benchmark comprises 96 sequences and 21,439 frames, divided into 48 training sequences with 11,474 frames and 48 test sequences with 9,965 frames.
Image resolutions range from $446\times270$ to $1920\times1080$.
The detection split contains 14,956 training and 6,483 validation images, with YOLO annotations, while the tracking data use the MOTChallenge format with target identities.
The sequences include takeoff, landing, turning, and circling, with small targets, similar appearances, dense formations, and mutual occlusions complicating identity association.
These characteristics make UAVSwarmV2 suitable for evaluating the tracking transfer of WorldUAV-trained detectors in Sec.~\ref{sec:uav_tracking}, including their effect on localization and identity consistency in real UAV swarms.

\section{Details on Synthetic Datasets}
\label{sec:appendix_synthetic_datasets}

Complementing Sec.~\ref{sec:related_synthetic_data}, this appendix describes synthetic datasets and data-generation methods based on simulation, image compositing, and learned generation.
We summarize their target tasks, synthesis mechanisms, and supervision, spanning image-level recognition and segmentation as well as video-based perception.
For generation frameworks and augmentation methods, we describe how training examples and annotations are constructed rather than treating them as fixed dataset releases.

\vspace{0.2cm}
\noindent\textbf{SYNTHIA.}
SYNTHIA~\cite{ros2016synthia} provides synthetic urban imagery for semantic segmentation.
A virtual city is used to render diverse street scenes and automatically assign pixel-level semantic labels, avoiding manual annotation of each generated image.
The collection captures variations in urban layout and observation conditions, allowing synthetic samples to supplement real-world segmentation data.
Its supervision is tied to the rendered scene, illustrating how a simulator can jointly produce images and dense labels for perception tasks.

\vspace{0.2cm}
\noindent\textbf{Kubric.}
Kubric~\cite{greff2022kubric} is a procedural data-generation framework that combines Blender rendering with PyBullet physics.
It produces scenes and videos with annotations derived from the underlying simulation, including object masks, depth, and motion information.
The released Multi-Object Video (MOVi) datasets vary object appearance, scene complexity, and camera motion to study object discovery from video.
By maintaining access to scene geometry and object states, Kubric supports consistent supervision across frames and offers a reusable infrastructure for constructing task-specific synthetic datasets.

\vspace{0.2cm}
\noindent\textbf{SynthText.}
SynthText~\cite{gupta2016synthetic} contains approximately 800,000 synthetic scene-text images with around eight million word instances.
Text is inserted into natural images using scene layout and geometric cues to produce plausible placement and orientation.
The annotations include transcriptions and bounding boxes at both word and character levels.
The collection demonstrates how real backgrounds can be combined with automatically labeled foreground content to scale training data while retaining much of the visual complexity of natural scenes.

\vspace{0.2cm}
\noindent\textbf{Copy-Paste.}
Copy-Paste~\cite{ghiasi2021simple} constructs augmented instance-segmentation examples by transferring segmented objects between training images.
Random object placement creates new foreground--background combinations, while the instance masks provide supervision for the resulting compositions.
The method improves data efficiency and recognition of rare categories without requiring an explicit scene simulator.
It is an image-level augmentation procedure rather than a fixed synthetic benchmark, and its original formulation does not prescribe object trajectories or temporal consistency across a video.

\vspace{0.2cm}
\noindent\textbf{Dataset Diffusion.}
Dataset Diffusion~\cite{nguyen2023dataset} generates synthetic images and semantic-segmentation pseudolabels using Stable Diffusion.
Text prompts specify the desired categories, while cross-attention and self-attention maps are used to infer their spatial extent.
Uncertain regions are excluded from the segmentation loss to reduce the effect of imperfect generated labels.
Evaluations on PASCAL VOC and MS COCO show how a pretrained image generator can provide dense training supervision without a graphics engine, although the labels remain estimates derived from model attention.

\vspace{0.2cm}
\noindent\textbf{TrackDiffusion.}
TrackDiffusion~\cite{li2025trackdiffusion} generates videos conditioned on object tracklets, allowing prescribed trajectories to guide object motion across frames.
Instance-aware conditioning and a temporal instance enhancer help maintain object consistency as targets move and change scale.
The generated sequences can augment training data for tracking models, connecting controllable video generation with downstream temporal perception.
This approach extends data synthesis from independent images to sequences whose spatial conditions explicitly encode the evolution of individual targets.

\vspace{0.2cm}
\noindent\textbf{Cosmos.}
Cosmos~\cite{nvidia2025cosmos} is a world foundation model platform for physical AI, comprising video-data curation, video tokenization, pretrained generative models, and examples of task-specific adaptation.
Its relevance to data synthesis lies in learning visual and temporal regularities from video and using them to generate conditioned scene evolution.
The platform provides a foundation for constructing application-specific video generators; it is not itself a UAV detection dataset with a prescribed set of bounding-box annotations.
This distinction motivates the explicit trajectory conditioning and annotation process used to construct WorldUAV.

% \clearpage
\section{Dataset Structure and Annotations}
\label{sec:appendix_dataset_structure}

\begin{figure}[t]
    \centering
    \includegraphics[width=\linewidth]{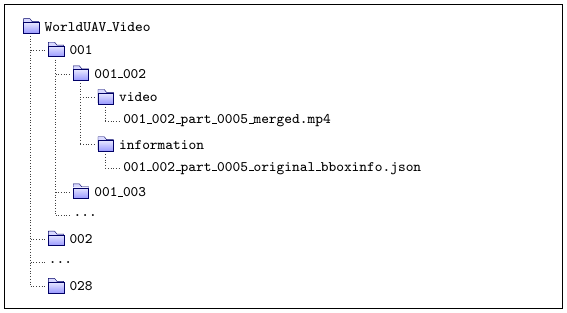}
    \caption{Folder organization of WorldUAV\_Video. Each illustrated sequence directory contains synthesized videos and corresponding JSON annotations, which record video attributes and frame-level bounding boxes. Representative entries are expanded, while repeated directory structures are omitted.}
    \label{fig:worlduav_video_structure}
    \vspace{-0.2cm}
\end{figure}

\begin{figure}[t]
    \centering
    \includegraphics[width=\linewidth]{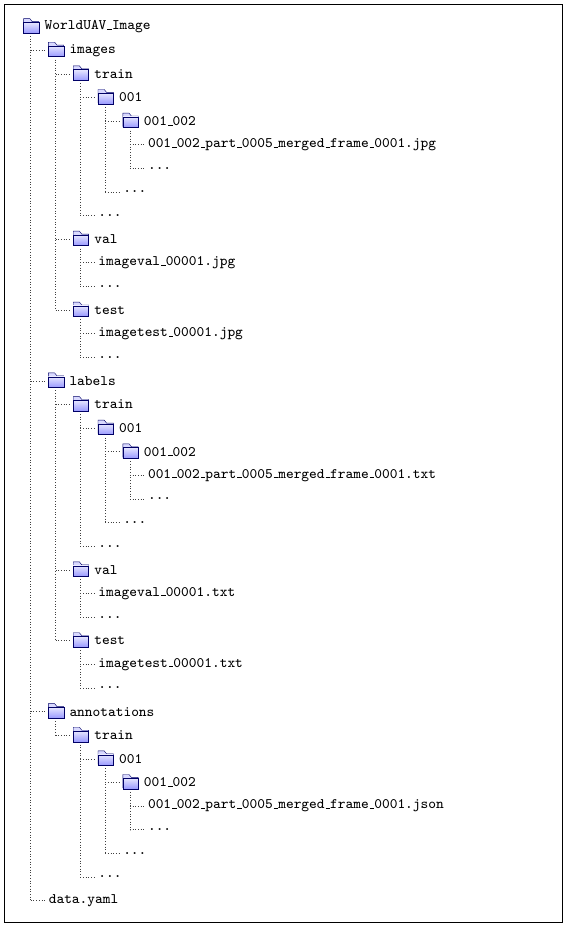}
    \caption{Folder organization of WorldUAV\_Image. Training images and YOLO labels retain the nested directory structure, while validation and test files are stored directly in their respective split directories with matching image and label names. The annotations directory contains additional JSON metadata for training images. }
    \label{fig:worlduav_image_structure}
\end{figure}

WorldUAV is organized into two complementary collections: \texttt{WorldUAV\_Video}, which stores synthesized video clips and their temporal annotations, and \texttt{WorldUAV\_Image}, which provides selected frames and detection labels.
Fig.~\ref{fig:worlduav_video_structure} and~\ref{fig:worlduav_image_structure} illustrate their respective directory structures.
In addition to bounding boxes, the annotation schema records the synthesis prompt, environmental conditions, UAV attributes, and source-frame information, supporting both spatial localization and analysis of the conditions under which each sample was generated.

\vspace{0.2cm}
\noindent\textbf{Video Organization.}
The video collection contains 6,878 MP4 clips, organized under numeric group and source-video directories, such as \texttt{001/001\_002}.
Each source directory contains a \texttt{video} folder for synthesized clips and an \texttt{information} folder for the corresponding JSON records.
Each video JSON record includes \texttt{video\_name}, \texttt{video\_path}, \texttt{predicted\_positions}, and \texttt{video\_resolution}.
The \texttt{video\_path} field identifies the synthesized clip, while \texttt{predicted\_positions} stores a sequence of records containing a frame index and its bounding box.
These temporal annotations retain the planned UAV trajectory after it has been mapped from the edited crop to the full video frame, as described in Sec.~\ref{sec:construction_pipeline}.

\vspace{0.2cm}
\noindent\textbf{Image Organization.}
The image collection contains 205,778 JPEG images, partitioned into 190,778 training, 5,000 validation, and 10,000 test images.
The \texttt{images} and \texttt{labels} directories share the same split structure, and each image is paired with a text label through its relative path and filename stem.
Training samples retain the nested group and source directories, and their filenames encode the parent clip and frame index.
Validation and test samples are stored directly under their split directories and are sequentially named \texttt{imageval\_00001} and \texttt{imagetest\_00001}, respectively, with matching \texttt{.jpg} and \texttt{.txt} extensions.

Additional image metadata are stored as JSON files under \texttt{annotations/train}, following the same relative paths and filename stems as the training images.
Each record includes the image \texttt{resolution}, parent \texttt{video\_path}, source \texttt{frame} index, and a \texttt{bbox}, together with the semantic attributes described below.
The root-level \texttt{data.yaml} provides the dataset configuration and UAV class definition.

\vspace{0.2cm}
\noindent\textbf{Semantic and Generation Metadata.}
The \texttt{prompt} field records the video prompt used for synthesis.
The \texttt{background\_scene}, \texttt{weather}, and \texttt{illumination} fields describe scene content and environmental conditions inherited from the source footage.
Their vocabularies comprise seven background categories, five weather conditions, and four lighting conditions, as detailed in Sec.~\ref{sec:dataset_coverage}.
The \texttt{drone\_type} and \texttt{drone\_pose} fields specify the UAV reference model and categorical viewpoint where available.
The latter refers to the viewpoint vocabulary defined in the main paper, rather than a continuous three-dimensional pose annotation.

% \clearpage
\section{Dataset Construction Settings}
\label{sec:appendix_construction_settings}

This section specifies the implementation settings and MLLM instructions for the four stages of the construction pipeline in Sec.~\ref{sec:construction_pipeline}. Fig.~\ref{fig:prompt_videorange}--\ref{fig:prompt_videoconsistency} reproduce the four English prompts in their order of use.

\subsection{Diverse Data Acquisition}
We automatically search Pixabay and Pexels for source videos covering diverse real-world environments. Retrieved footage is resampled to 25 fps and segmented into 125-frame clips, corresponding to five seconds of video, before scene assessment and synthesis.

\subsection{Scene-Aware Flight Planning}
\label{sec:appendix_planning_settings}
Planning comprises two sequential API calls to Doubao-Seed-2.0 Pro~\cite{bytedance2026seed2}, using model identifier \texttt{doubao-seed-2-0-pro-260215} and the API base URL \url{https://ark.cn-beijing.volces.com/api/v3}. For both calls, video preprocessing samples at 1 fps, equivalent to approximately one frame per 25 source frames (4\%) after frame-rate normalization.

\begin{figure*}[p]
    \centering
    \includegraphics[width=\textwidth]{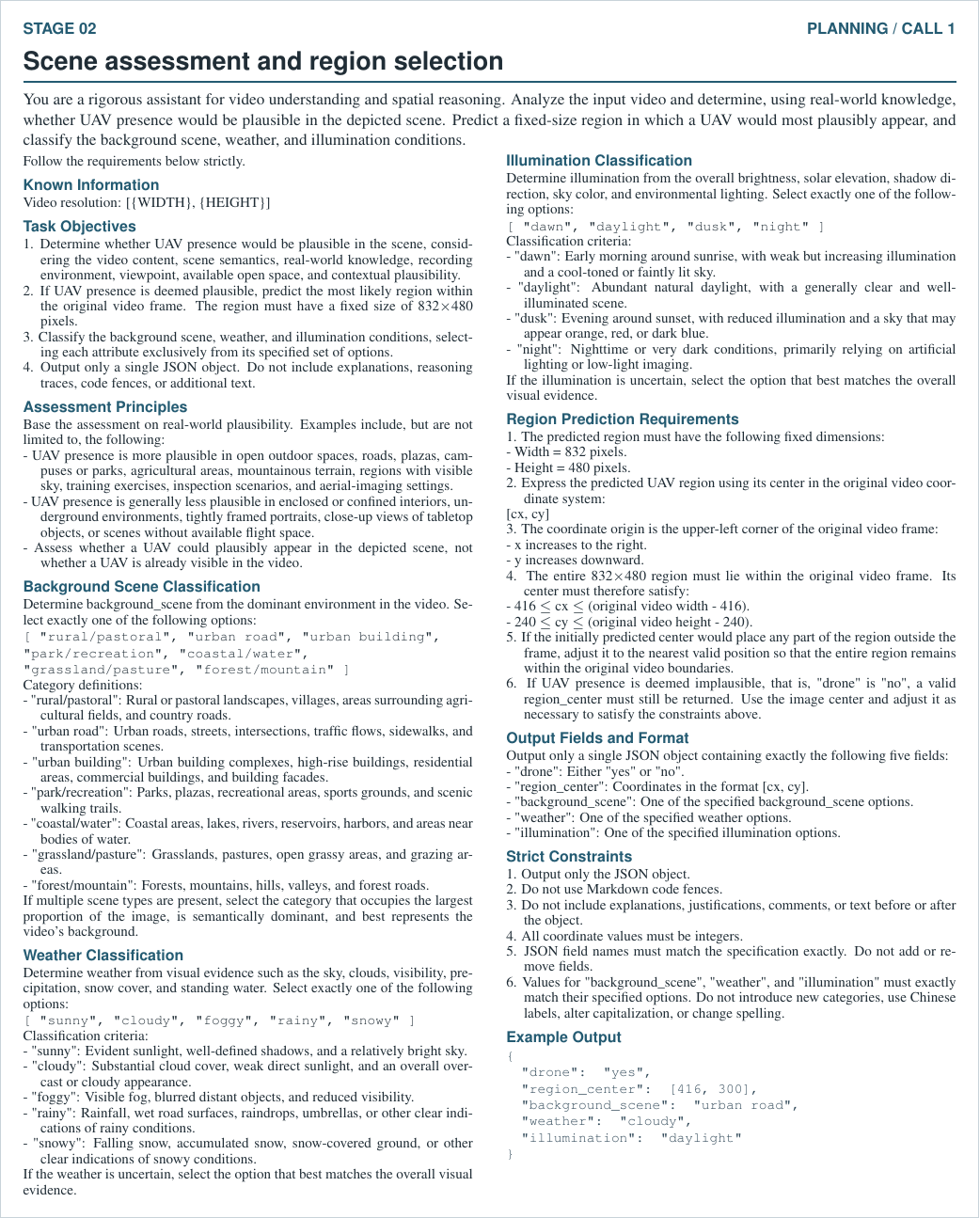}
    \caption{\textbf{Planning, call 1.} Complete prompt for scene assessment, region selection, and environmental annotation.}
    \label{fig:prompt_videorange}
\end{figure*}

\noindent\textbf{Call 1: Scene Assessment and Region Selection.}
The first call evaluates whether UAV presence is plausible, selects a fixed $832\times480$ region within the original frame, and assigns background, weather, and illumination categories (Fig.~\ref{fig:prompt_videorange}). The resolution placeholders in the prompt are populated with the source-frame dimensions. Clips assessed as unsuitable are excluded; the selected region provides the video input to the next call.

\begin{figure*}[p]
    \centering
    \includegraphics[width=\textwidth]{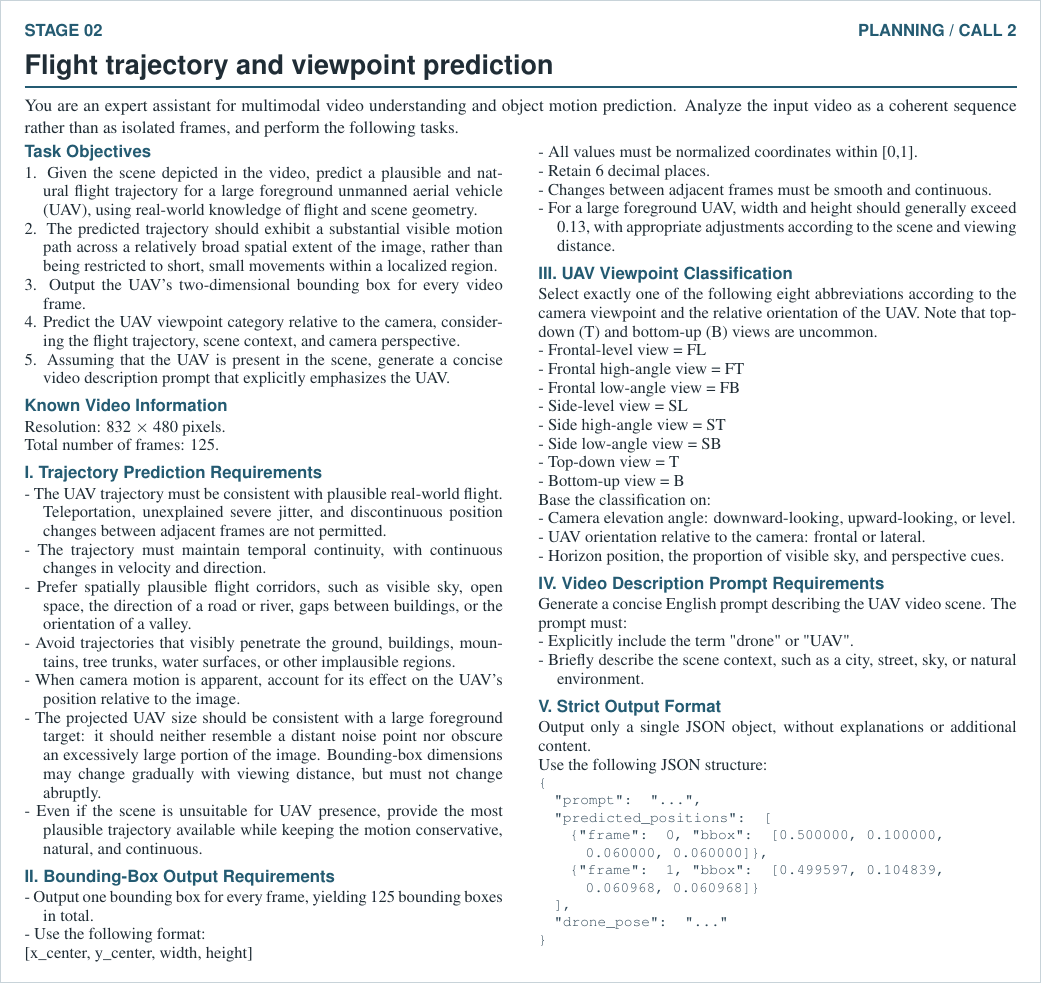}
    \caption{\textbf{Planning, call 2.} Complete prompt for trajectory prediction, viewpoint classification, and synthesis-prompt construction. The JSON example shows two entries; the instructions require 125 frame-level boxes.}
    \label{fig:prompt_dronepredicted}
\end{figure*}

\noindent\textbf{Call 2: Flight Trajectory and Viewpoint Prediction.}
Given the selected crop, the second call produces a synthesis prompt, a UAV viewpoint category, and 125 normalized bounding boxes, each represented as $(x_c,y_c,w,h)$ (Fig.~\ref{fig:prompt_dronepredicted}). These boxes specify the intended position and scale in every frame. The instructions encourage a spatially extended, continuous flight path that respects scene geometry and camera motion. The dense trajectory is predicted from the sampled observations; it is not a set of independently observed frame-level detections.

\begin{figure}[t]
    \centering
    \includegraphics[width=0.48\textwidth]{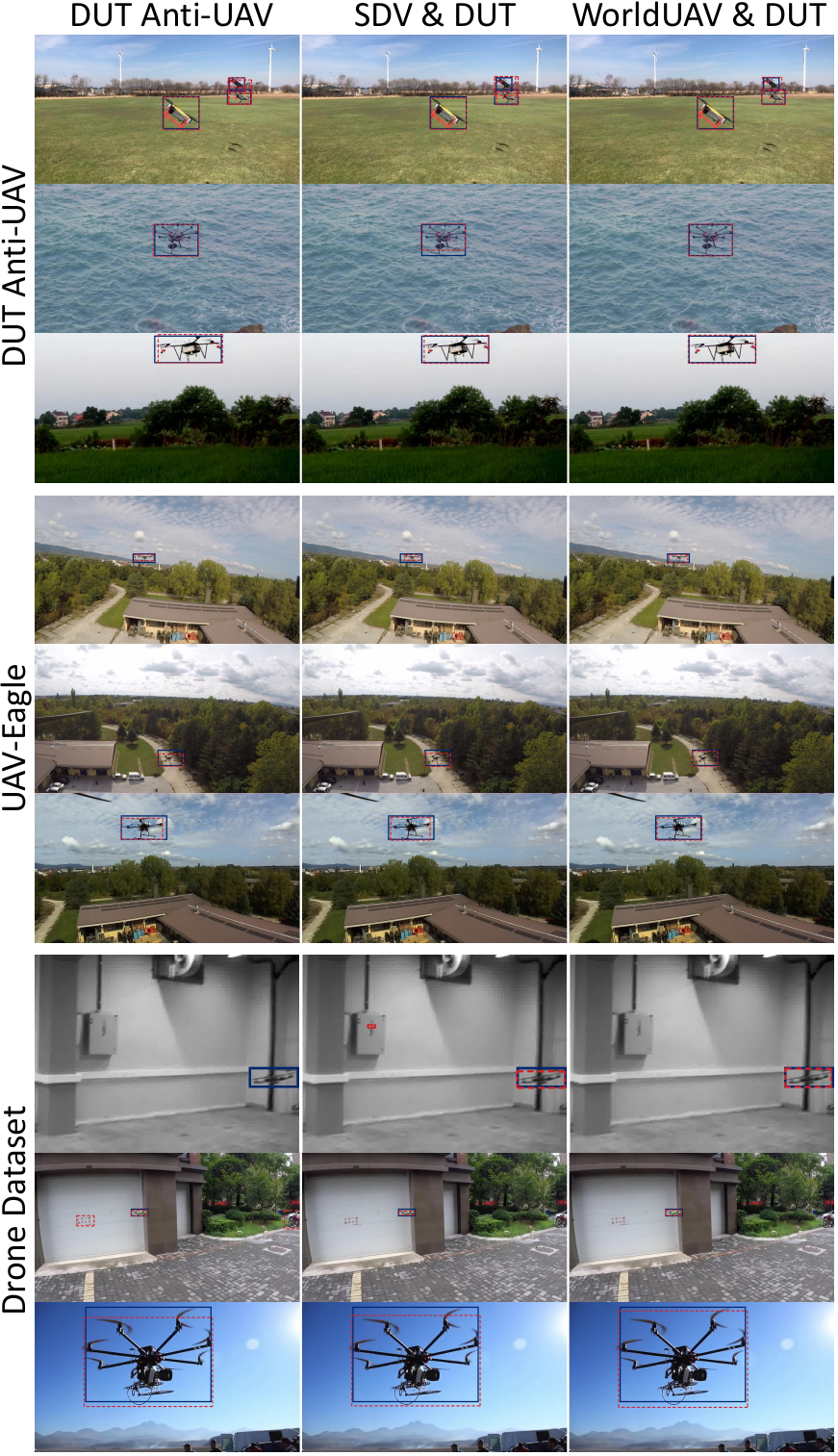}
    \caption{\textbf{Qualitative comparison of YOLOv12l detections.}
    Columns from left to right correspond to training on DUT, SDV \& DUT, and WorldUAV \& DUT.
    The three row groups show DUT Anti-UAV, UAV-Eagle, and Drone Dataset, from top to bottom.
    Each row displays the same test image under all three training configurations.
    Ground-truth boxes are drawn with dark-blue solid lines, and predictions with red dashed lines.}
    \label{fig:worlduav_detection_examples}
\end{figure}

\subsection{World Model-Driven UAV Synthesis}
We deploy HunyuanCustom~\cite{hu2025hunyuancustom} locally for inference on an NVIDIA H100 GPU. Each invocation conditions on the $832\times480$ video crop, a UAV reference image, the synthesis prompt, and binary masks rasterized from the planned boxes. We set \texttt{--seed} to \texttt{1024}, \texttt{--infer-steps} to \texttt{50}, and \texttt{--flow-shift-eval-video} to \texttt{5.0}, fixing the random seed, the number of denoising steps, and the flow-matching timestep shift, respectively.

The \texttt{--expand-scale} parameter controls morphological expansion or contraction of the input masks. We set it to \texttt{0}, preserving the planned mask extent without dilation or erosion. This setting is intended to let the UAV occupy the designated mask region without introducing an additional spatial margin, promoting tight correspondence between the synthesized target and its initial bounding box. The generated crop is composited into the original frames, and the planned boxes are transformed to full-frame coordinates to initialize the annotations.

\subsection{Spatiotemporal Quality Verification}
Verification consists of two sequential MLLM calls using the same API endpoint, model, and video sampling configuration as Sec.~\ref{sec:appendix_planning_settings}. The frame-based recognizability check supplies images directly, as specified below, rather than relying on the video preprocessing sampler. A clip is retained only if it passes both checks.

\begin{figure*}[p]
    \centering
    \includegraphics[width=0.94\textwidth]{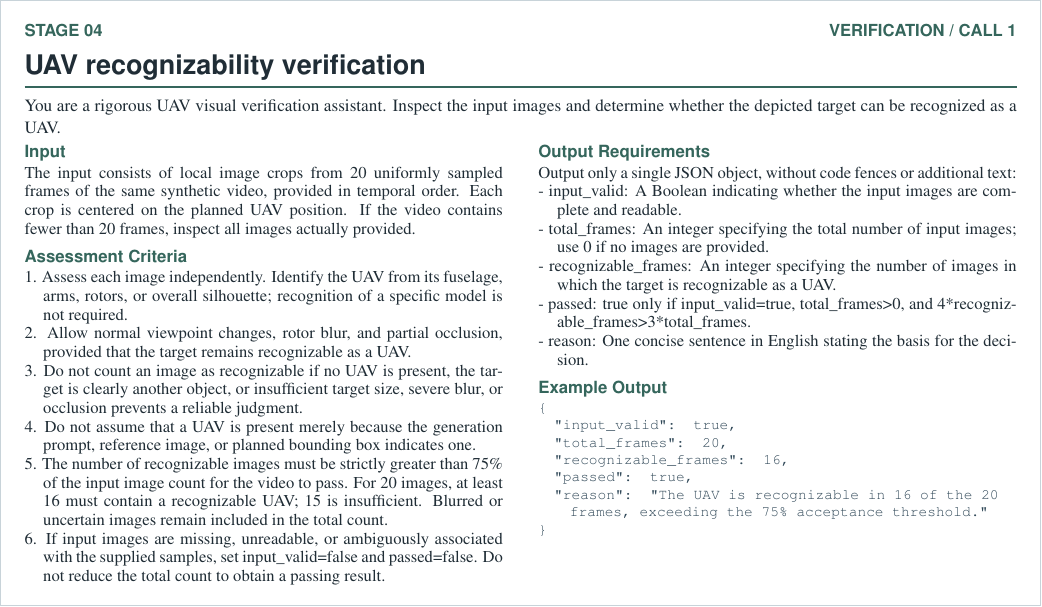}
    \caption{\textbf{Verification, call 1.} Complete prompt for UAV recognizability assessment from uniformly sampled local crops.}
    \label{fig:prompt_droneverification}
    \vspace{6pt}
    \includegraphics[width=0.94\textwidth]{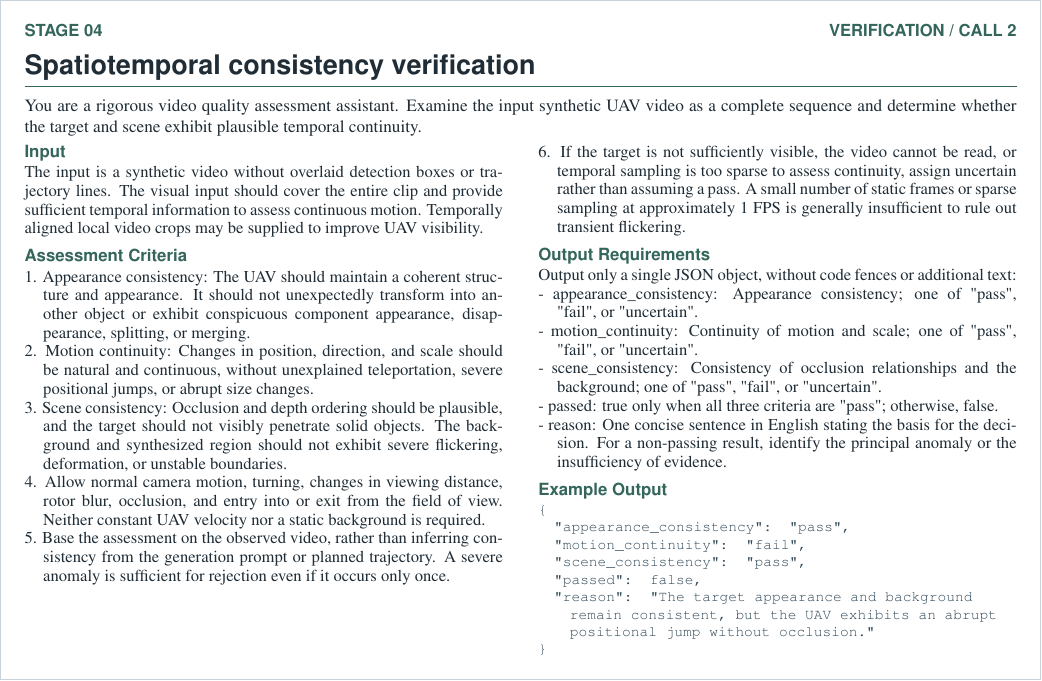}
    \caption{\textbf{Verification, call 2.} Complete prompt for appearance, motion, and scene consistency assessment, with explicit rejection when evidence is insufficient.}
    \label{fig:prompt_videoconsistency}
\end{figure*}

\noindent\textbf{Call 1: UAV Recognizability.}
The first call inspects local crops centered on the planned UAV positions in 20 uniformly sampled frames, supplied in temporal order (Fig.~\ref{fig:prompt_droneverification}). A clip passes only when more than 75\% of the inspected crops contain a recognizable UAV, requiring at least 16 of 20 crops. If fewer than 20 frames are available, all available frames are assessed. Uncertain crops remain in the denominator, and invalid inputs cannot pass.

\noindent\textbf{Call 2: Video Consistency.}
The second call evaluates the synthesized video without annotation overlays, checking UAV appearance, motion continuity, and consistency of occlusion and background content (Fig.~\ref{fig:prompt_videoconsistency}). Each criterion receives a \texttt{pass}, \texttt{fail}, or \texttt{uncertain} decision, and all three must pass. 

\section{Detection Examples}
\label{sec:appendix_detection_examples}

Fig.~\ref{fig:worlduav_detection_examples} presents selected detection examples from DUT Anti-UAV~\cite{zhao2022vision}, UAV-Eagle~\cite{barivsic2022brain}, and Drone Dataset~\cite{aksoy2019drone}.
All three configurations use YOLOv12l, trained on DUT alone, SynDroneVision (SDV)~\cite{lenhard2025syndronevision} combined with DUT, or WorldUAV combined with DUT, respectively.
The same image is repeated across each row to enable direct comparison of predicted boxes against the ground truth.

% Full-width pages keep the sampled video frames legible.
\onecolumn
\section{Randomly Sampled Videos from WorldUAV}
\label{sec:appendix_video_samples}

\begin{figure}[!ht]
    \centering
    \includegraphics[width=0.94\textwidth,pagebox=cropbox]{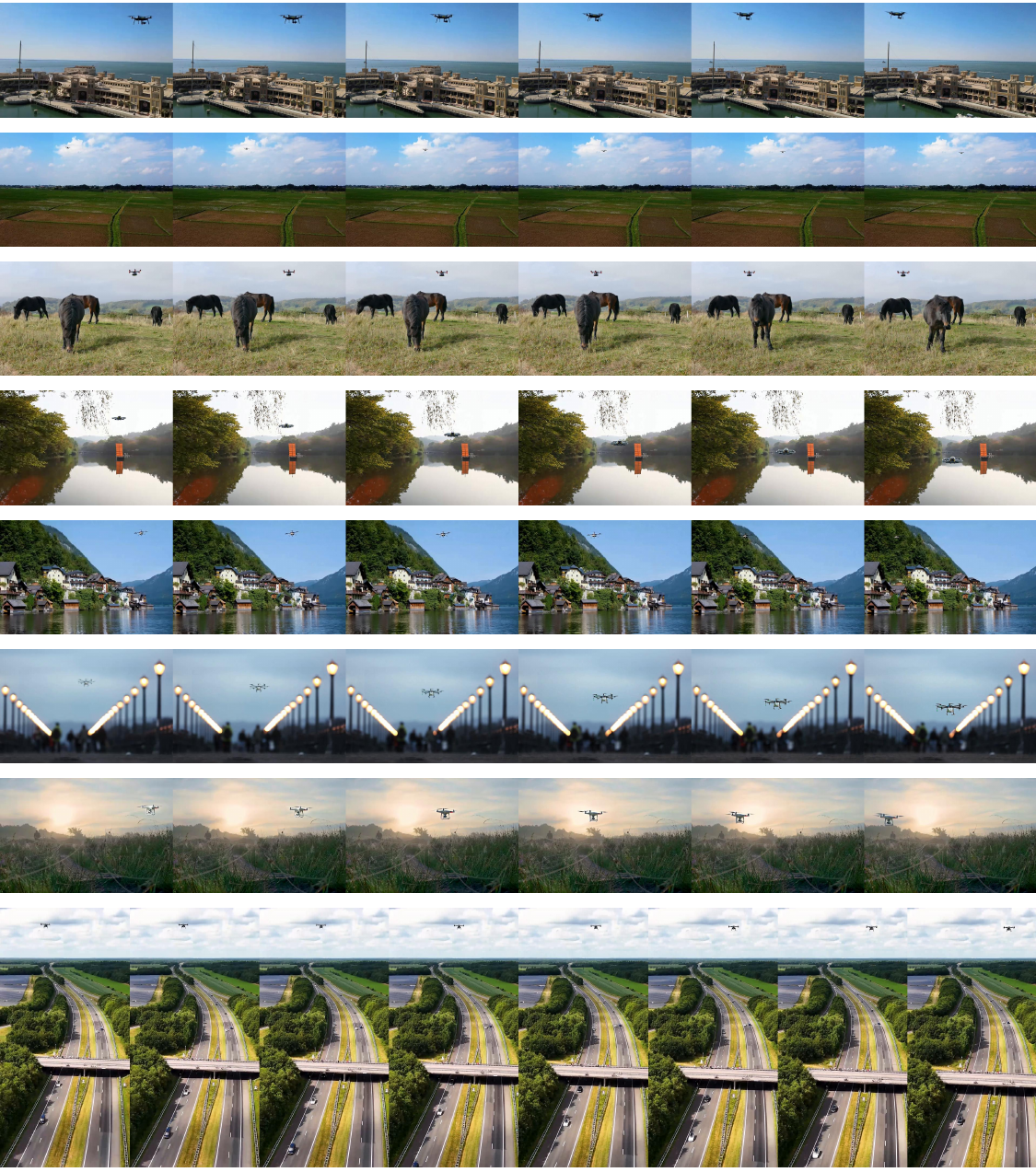}
    \caption{\textbf{Randomly sampled WorldUAV videos (1/3).} Each row corresponds to a single video, with sampled frames arranged chronologically from left to right.}
    \label{fig:worlduav_video_samples_1}
\end{figure}

\begin{figure}[!t]
    \centering
    \includegraphics[width=0.94\textwidth,pagebox=cropbox]{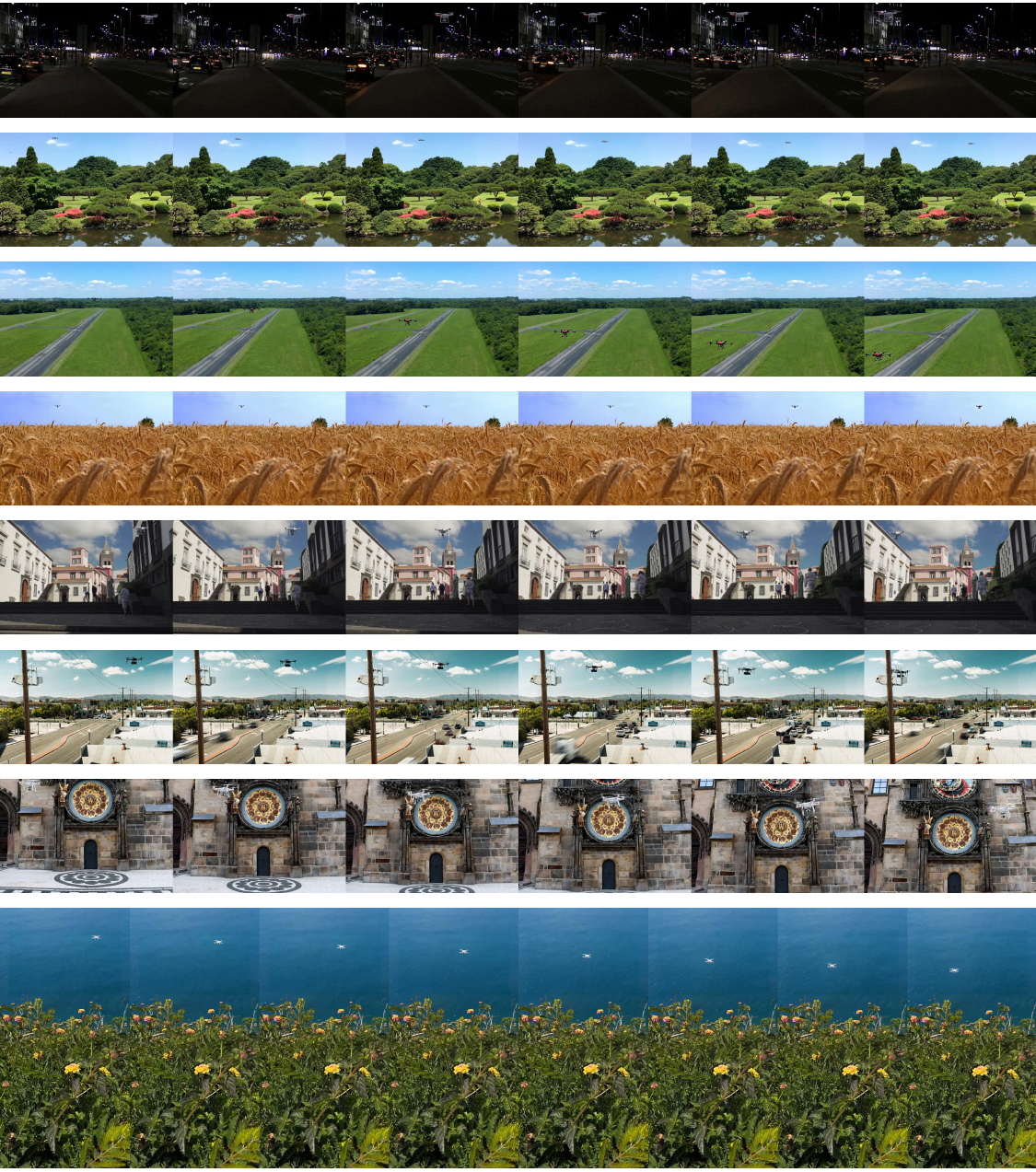}
    \caption{\textbf{Randomly sampled WorldUAV videos (2/3).} Each row corresponds to a single video, with sampled frames arranged chronologically from left to right.}
    \label{fig:worlduav_video_samples_2}
\end{figure}

\begin{figure}[!t]
    \centering
    \includegraphics[width=0.94\textwidth,pagebox=cropbox]{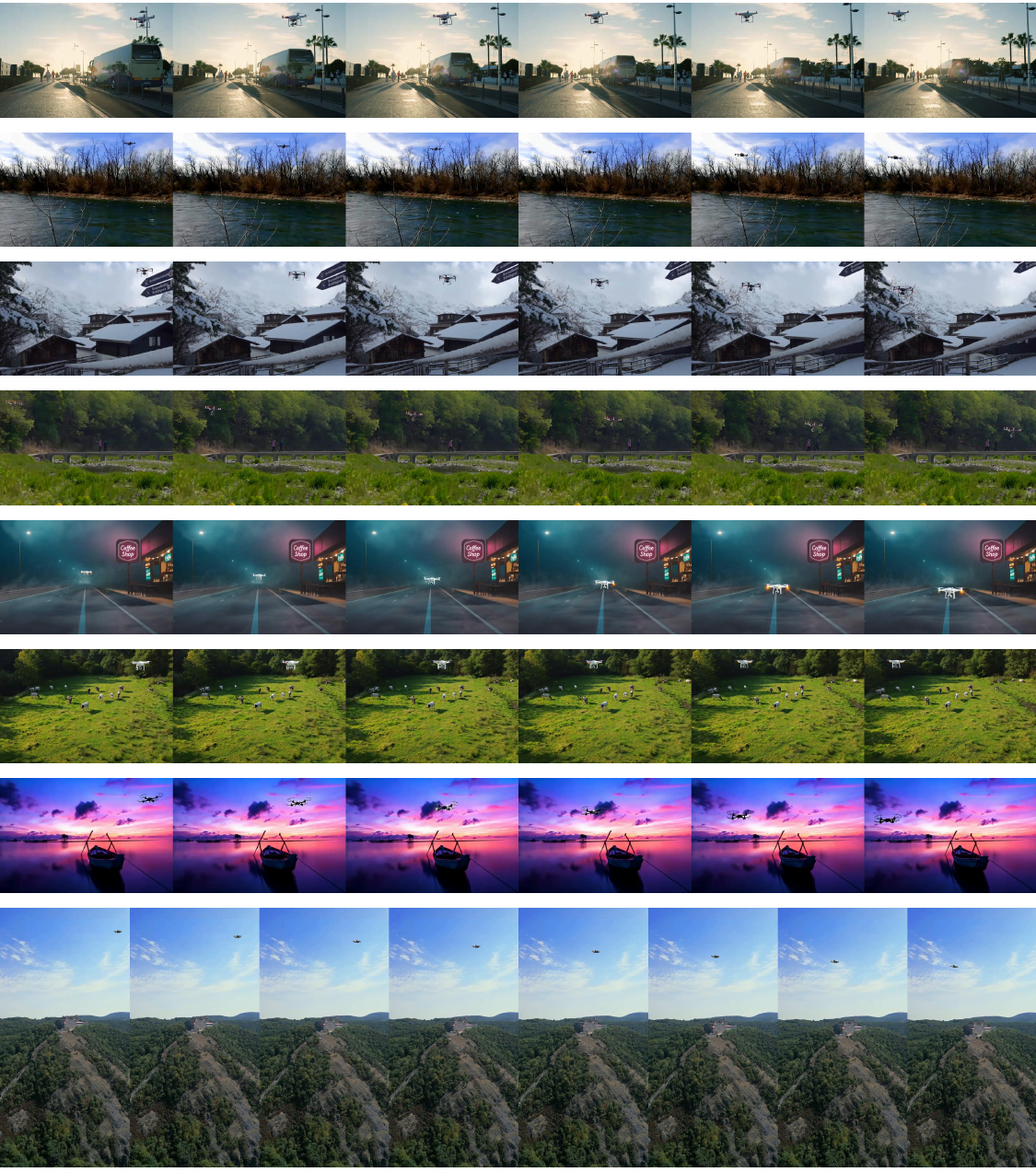}
    \caption{\textbf{Randomly sampled WorldUAV videos (3/3).} Each row corresponds to a single video, with sampled frames arranged chronologically from left to right.}
    \label{fig:worlduav_video_samples_3}
\end{figure}

Fig.~\ref{fig:worlduav_video_samples_1}--\ref{fig:worlduav_video_samples_3} present randomly sampled clips from \texttt{WorldUAV\_Video}, illustrating variation in UAV appearance, scene context, and environmental conditions.

\clearpage

\clearpage

% Restore the paper's layout for any subsequent appendix sections.
\twocolumn

\end{document}